\documentclass[12pt]{article}
\usepackage{graphicx,psfrag,epsf}
\usepackage{enumerate}

\newcommand{\blind}{0}

\usepackage{natbib}
\usepackage{aas_macros}
\usepackage{algorithm}
\usepackage[margin=0pt,font=small]{caption}
\usepackage{siunitx}
\usepackage{relsize}
\usepackage{amsmath}
\usepackage{graphicx}
\DeclareMathOperator*{\argmin}{arg\,min} 
\usepackage{amssymb}
\DeclareMathOperator{\E}{\mathbb{E}}
\usepackage{mathtools}
\usepackage{algorithmic}
\newcommand{\be}{\begin{equation}}
\newcommand{\ee}{\end{equation}}

\def\bi#1{\hbox{\boldmath{$#1$}}}

\begin{document}

\def\spacingset#1{\renewcommand{\baselinestretch}%
{#1}\small\normalsize} \spacingset{1}

\if0\blind
{
  \title{\bf Posterior inference unchained with EL$_2$O}
  \author{
    Uro\v s Seljak $^{1}$ $^{2}$, Byeonghee Yu $^{1}$ \\
    \small $^{1}$ Berkeley Center for Cosmological Physics and Department of Physics, University of California, \\ \small Berkeley, CA 94720, USA \\
    \small $^{2}$ Physics Division, Lawrence Berkeley National Laboratory, 1 Cyclotron Rd, \\ \small  Berkeley, CA 93720, USA}
  \date{\vspace{-5ex}}
  \maketitle
} \fi

\if1\blind
{
  \begin{center}
    {\LARGE\bf Title}
\end{center}
  \medskip
} \fi






\bigskip
\begin{abstract}
Statistical inference of analytically non-tractable posteriors is a difficult problem because
of marginalization of correlated variables and stochastic methods such as MCMC and VI are commonly used. We argue that KL divergence minimization used by MCMC and stochastic VI is based on stochastic integration, which is noisy. We propose instead $t(\ln t)^2$ f-divergence, which evaluates expectation of L$_2$ distance squared between the approximate log posterior q and the un-normalized log posterior of p, both evaluated at the 
same sampling point, and optimizes on $q$ parameters. If $q$ is a good approximation to $p$ it has to agree with it on every sampling point and because of this  the method we call EL$_2$O is free of sampling noise and has better optimization properties. As a consequence, increasing the expressivity of $q$ with
point-wise nonlinear transformations and Gaussian mixtures improves both the quality of results and the convergence rate. 
We develop Hessian, gradient and gradient free versions of the method, which can determine M(M+2)/2+1, M+1 and 1 parameter(s) of $q$ with a single sample, respectively. EL$_2$O value provides a reliable estimate of the quality of the approximating posterior.
We test it on several examples, including a realistic 13 dimensional galaxy clustering analysis, showing that it is 3 orders of magnitude faster than MCMC and 2 orders of 
magnitude faster than ADVI, while giving smooth and accurate non-Gaussian posteriors, often requiring dozens of iterations only.

\end{abstract}

\noindent%

\spacingset{1.45}
\section{Introduction}

The goal of statistical inference from data 
can be stated as follows: given some data determine 
the posteriors of some parameters, while marginalizing over other parameters. The 
posteriors are best parametrized in terms of 1D probability density 
distributions, but alternative 
descriptions such as the mean and the variance can sometimes be used, especially in the asymptotic 
regime of large data where they fully describe the posterior. Occasionally we also want 
to examine higher dimensional posteriors, such as 2D probability density 
plots, but we rarely 
go to higher dimensions due to the difficulty of visualizing it. While we want to 
summarize the results in a series of 1D and 2D plots, the actual problem can have a 
large number of dimensions, most of which we may not care about, but which 
are correlated with the ones we do. 
The main difficulty in obtaining reliable lower dimensional posteriors
lies in the marginalization part: marginals, i.e., averaging over the probability 
distribution of certain parameters, can change the answer significantly
relative to the answer for the unmarginalized 
posterior, where those parameters are assumed to 
be known. 

A standard approach to posteriors is Monte Carlo Markov Chain (MCMC) sampling.
In this approach we sample all the parameters according to their probability 
density. After the samples are created marginalization is trivial, as one 
can simply count the 1D and 2D posterior density 
distribution of the parameter of interest. MCMC is 
argued to be exact, in the sense that in the limit of 
large samples it converges to the true answer. But in practice
this limit may never be reached. For example, 
doing Metropolis sampling without knowing the covariance structure 
of the variables suffers from the curse of dimensionality. 
In very high dimensions this is practically impossible. 
One avoids the curse of 
dimensionality by having access to the gradient of the loss 
function, as in Hamiltonian 
Monte Carlo (HMC). 
However, in high dimensions
thousands of model
evaluations may be needed to produce a single independent sample (e.g. \cite{JascheWandelt13}), 
which often makes it prohibitively expensive, specially if model evaluation 
is costly. 

Alternatives to MCMC are approximate methods such as Maximum Likelihood Estimation (MLE) 
or its Bayesian version, Maximum A Posteriori (MAP) estimation,
or KL divergence minimization based variational inference (VI). These can be less expensive, 
but can also give inaccurate results and must be used with care. MAP can give incorrect 
estimators in many different situations, even in the limit of large data, and 
is thus an inconsistent estimator, a result known 
since \cite{NeymanScott1948}. Similarly, mean field VI can give an incorrect 
mean in 
certain situations. Full rank VI (FRVI) typically gives 
the posterior mean close to the correct value, but not always 
(we present one example 
in section \ref{sec4}). It often does not give correct variance.
Quantifying the error of the approximation is difficult \citep{YaoVSG18}.
A related method is Population Monte Carlo (PMC) 
\citep{CappeDoucEtAl08,WraithKilbingerEtAl09}, which uses sampling 
from a proposal distribution to obtain new samples, and the posterior at the 
sample to improve upon the proposal sampling density. Both of these methods 
use KL divergence to quantify the agreement between the proposal 
distribution and the true distribution. 

In many scientific applications the cost of evaluating the model and 
the likelihood
can be very high. In these situations MCMC becomes prohibitively 
expensive. 
The goal of this paper is to develop a method that is optimization 
based and 
extends MAP and stochastic VI methods such as ADVI \citep{KucukelbirTRGB17}, such that it requires a 
low number 
of likelihood evaluations, while striving to be as accurate as MCMC.\footnote{ In general exact inference is impossible because global optimization of non-convex surfaces is an unsolved problem in high dimensions.} We would like to 
avoid some of the main pitfalls of the approximate 
methods like MAP or VI. 
Our goal is to have a method that works for both 
convex and non-convex problems, and works for moderately high 
dimensions, where a full rank matrix inversion is not a computational 
bottleneck: this can be a dozen or up to a few thousand dimensions, 
depending on the computational cost of the likelihood
and the complexity of posterior surface. 

The outline of the paper is as follows. 
In section \ref{sec2} we compare traditional stochastic KL divergence 
approaches to our new proposal of using L$_2$ distance on a 
toy 1D Gaussian example. 
In section \ref{sec3} we develop the method further by 
incorporating higher derivative information, and increasing 
the expressivity of approximate posteriors, while still allowing 
for analytic marginalization. In section \ref{sec4} we show 
several examples, including a realistic data analysis example 
from our research area of cosmology. We conclude by discussion
and conclusions in section \ref{sec5}. 

\section{Stochastic KL divergence minimization versus ${\bf EL_2O}$}
\label{sec2}

A general problem of statistical inference is how to infer parameters from the data:
we have some data $\bi{x}= \{x_i\}_{i=1}^N$ and some parameters
the data depend on, $\bi{z}= \{z_j\}_{j=1}^M$. 
We want to 
describe the posterior of $\bi{z}$ given data $\bi{x}$. 
We can define the posterior $p(\bi{z}|\bi{x})$ as
\begin{equation}
p(\bi{z}|\bi{x})=\frac{p(\bi{x}|\bi{z})p(\bi{z})}{p(\bi{x})}=\frac{p(\bi{x},\bi{z})}{p(\bi{x})},
\label{loss}
\end{equation}
where $p(\bi{x}|\bi{z})$ is the likelihood of the data, $p(\bi{z})$ is the prior of $\bi{z}$
and $p(\bi{x})=\int p(\bi{x}|\bi{z})p(\bi{z})d\bi{z}$ is the normalization. 
In general we have access to the prior and likelihood, but not the normalization. 
We can define the negative log of posterior in terms of what we have access to, which is negative log joint distribution $\tilde{\mathcal{L}}_p$, defined as 
\begin{equation}
\tilde{\mathcal{L}}_p=-\ln p(\bi{x},\bi{z})=-\ln p(\bi{x}|\bi{z})-\ln p(\bi{z})= -\ln p(\bi{z}|\bi{x})-\ln p(\bi{x})\equiv \mathcal{L}_p-\ln p(\bi{x}).
\label{loss1}
\end{equation}
For flat prior this is simply the negative log likelihood of the data. 
Note 
that the difference between $\tilde{\mathcal{L}}_p$ and $\mathcal{L}_p$
is $\ln p(\bi{x})$, which is independent of $\bi{z}$, so in terms 
of gradients with respect to $\bi{z}$ there is no difference between 
the two and we will not distinguish between them. 


We would like to have accurate posteriors, but we would also like to avoid 
the computational cost of MCMC. 
Our goal is to describe the posteriors of parameters, and our approach will be rooted in 
optimization methods such as MAP or VI, where we assume a simple analytic form for the posterior, and try 
to fit its parameters to the information we have. 

To explain the motivation behind our 
approach we will for simplicity in this section 
assume we only have a single parameter 
$z$ given the data $\bi{x}$, 
  $\tilde{\mathcal{L}}_p=-\ln p(z|\bi{x})+\ln p(\bi{x})$.
We would like to fit the posterior of $z$ to a simple form, and the Gaussian ansatz is the 
simplest, 
\begin{equation}
\mathcal{L}_q=-\ln q(z), \, q(z)=N(z;\mu,\Sigma).
\label{1drank}
\end{equation}
We will also assume the posterior is Gaussian for the purpose 
of expectations, but since this is something we do not know a 
priori we will perform the estimation of parameters of $q$. 

\subsection{Stochastic KL divergence minimization}

Many of the most popular statistical inference methods are rooted in the minimization of 
KL divergence, defined as 
\begin{equation}
{\rm KL}(q||p)=\E_q(\mathcal{L}_p-\mathcal{L}_q), \, {\rm KL}(p||q)= \E_p(\mathcal{L}_q-\mathcal{L}_p). 
\end{equation}
Here $\E_{q}$, $\E_{p}$ is the expectation over $q$ and $p$, respectively. 

For intractable posteriors this cannot be evaluated exactly, and one tries to minimize 
KL divergence sampled over the corresponding probability distributions. Deterministic evaluation using quadratures
is possible in 
very low dimensions or under the mean field assumption, but this
becomes impossible in more than a few dimensions and we will not 
consider it further here. 
In this case stochastic sampling is the only practical 
method: we will thus consider stochastic minimization of KL divergence. 

We first briefly show that stochastic 
minimization of ${\rm KL}(p||q)$ is noisy (this is a known result and 
not required for the rest of the paper). 
Let's assume 
we have 
generated $N_k$ samples $z_k$ from $p$, using for example MCMC, with which we evaluate $\mathcal{L}_p(z_k)$ for $k=1,..N_k$. 
 We have 
\begin{equation}
{\rm KL}(p||q)=N_k^{-1}\left[\sum_k -\mathcal{L}_p(z_k) + \frac{(z_k-\mu)^2}{2\Sigma}+\frac{\ln (2\pi \Sigma)}{2}\right]. 
\end{equation}
Let us minimize 
${\rm KL}(p||q)$ with respect to $\mu$ and $\Sigma$. We find that $\mathcal{L}_p(z_k)$ do not 
enter into the answer at all, and we get 
\begin{equation}
\mu=N_k^{-1}\sum_k z_k,\, \Sigma=N_k^{-1}\sum_k (z_k-\mu)^2.
\end{equation}
As expected 
this is the standard Monte Carlo (MC) result for the 
first two moments of the posterior given the MCMC samples
from $p$. The answer converges to the true value as 
$N_k^{-1/2}$: one requires many samples for convergence.
MCMC sampling usually requires many calls to $\tilde{\mathcal{L}}_p(z)$ before an independent sample 
of $z_k$ is generated, with the correlation length strongly dependent on the nature of the 
problem and the sampling method. 
If one instead tries to approximate the
moments of $p$ one obtains Expectation 
Propagation method \citep{Minka01}.

Now let us look at stochastic minimization of ${\rm KL}(q||p)$. This corresponds to Variational Inference (VI)
\citep{WainwrightJordan08,BleiEtAl16}, which is argued to be significantly faster than MCMC. 
Here we assume $q$ is approximate posterior with a known 
analytic form, but since the posterior $p$ 
is not analytically 
tractable we will create samples from $q$ (an example of this 
procedure is ADVI, \cite{KucukelbirTRGB17}). 
Let us define the samples as 
\begin{equation}
z_k=\Sigma^{1/2}\epsilon_k+\mu,  
\label{zk}
\end{equation}
where $\epsilon_k$ is a random number drawn from a unit variance zero mean 
Gaussian $N(\epsilon_k;0,1)$. 
With this we find
\begin{equation}
{\rm KL}(q||p)=N_k^{-1}\sum_k\left[  -\frac{\epsilon_k^2}{2}-\frac{\ln (2\pi \Sigma)}{2}+\mathcal{L}_p(z_k)\right]. 
\end{equation}
We want to use $\mathcal{L}_p(z_k)$ to update information on
the mean $\mu$, but it only enters via $z_k$ inside 
$\mathcal{L}_p(z_k)$. 
So if we want to minimize KL divergence with respect to $\mu$ we have to propagate its 
derivative through $z_k$, the so called reparametrization trick \citep{KingmaWelling13,RezendeMW14}. 

To proceed let us assume that the posterior is given by a Gaussian
\begin{equation}
p(z|\bi{x})=N(z;\mu_t,\Sigma_t),
\end{equation}
where the subscripts $t$ denote true value. Since we are for now assuming that 
we do not have access to the analytic 
gradient, we will envision that the gradient with 
respect to the $\mu$ and $\Sigma$ parameters inside $z_k$ can be evaluated via a finite 
difference, evaluating $\nabla_{z}\mathcal{L}_p(z_k)=[\mathcal{L}_p(z_k+\delta z_k)-\mathcal{L}_p(z_k)]/\delta z_k)$. With $\nabla_{\mu}\mathcal{L}_p=\nabla_{z}\mathcal{L}_p (dz/d\mu)=\nabla_{z}\mathcal{L}_p$ we find that the gradient of KL divergence with 
respect to $\mu$ equal zero gives
\begin{equation}
\nabla_{\mu} {\rm KL}(q||p)=\sum_k \frac{(z_k-\mu_t)}{\Sigma_t}=0,\, \mu=\mu_t-N_k^{-1} \sum_k\Sigma^{1/2}\epsilon_k. 
\end{equation}
Since the mean of $\epsilon_k$ is zero this will converge to the correct answer, but 
will be noisy and the 
convergence to the true value will be as $N_k^{-1/2}$. 

To solve for the variance we similarly take a gradient with respect to $\Sigma$ and set it to zero,
\begin{equation}
\nabla_{\Sigma} {\rm KL}(q||p)=-\frac{1}{2\Sigma}+\sum_k \frac{\Sigma^{-1/2}\epsilon_k(z_k-\mu_t)}{2\Sigma_t}=0,
\end{equation}
with solution 
\begin{equation}
\Sigma=\frac{N_k\Sigma_t}{\sum_k \left[\epsilon_k^2+(\mu-\mu_t)\Sigma^{-1/2}\epsilon_k \right]}.
\end{equation}
Note that this is really a quadratic equation in $\Sigma^{1/2}$, which may have multiple roots: minimizing 
stochastic ${\rm KL}(q||p)$ is not a convex optimization problem. 
Even if we have converged on $\mu=\mu_t$ rapidly so that we can drop the last term in the 
denominator and avoid solving this as a quadratic equation, we are still left with a fluctuating term $(\sum_k\epsilon_k^2)^{-1}$. This expression
also converges as $N_k^{-1/2}$ to the true value. As we iterate 
towards the correct solution we also have to vary $q$, so the overall number of calls 
to $\tilde{\mathcal{L}}_p(z_k)$ will be larger. Results are shown in figure \ref{fig:ADVI}. 

In summary, minimizing ${\rm KL}(p||q)$ and ${\rm KL}(q||p)$ with sampling is a noisy
process, converging to the true answer with $N_k$ samples as $N_k^{-1/2}$. We will argue below this is a consequence of KL 
divergence integrand not being positive definite. 
In this context it is not immediately obvious 
why should stochastic VI be faster than MCMC, except that the prefactor for MCMC
is typically larger, because the MCMC 
samples of $p$ are correlated, while VI samples drawn from $q$ are not (this is
however somewhat 
offset by the fact that in stochastic VI one must also iterate on $q$). 

\subsection{EL$_2$O: Optimizing the expectation of L$_2$ distance squared of log posteriors}

Minimizing KL divergence is not the 
only way to match two probability distributions. 
Recent work has argued that 
KL divergence objective function may not be optimal, 
and that other
objective functions may have better 
convergence properties \citep{RanganathTAB16}. In this 
work we will also modify the objective function, but with the goal 
of preserving the expectation of 
KL divergence minimization in appropriate limits. 
A conceptually simple approach is to 
minimize the Euclidean distance squared between the true and approximate log posterior
averaged over the samples drawn from some fiducial posterior $\tilde p$ 
close to the posterior $p$: 
this too will be zero when the two 
are equal, and will be averaged over $\tilde p$: as long as $\tilde p$ is close to $p$ this will 
provide approximately correct weighting for the samples. 
If the distance is not zero it will also 
provide an estimate of the error generated by $q$ not being equal to $p$, which can be 
reduced by improving on $q$. While current 
estimate for $q$ can be used for $\tilde p$, and iterate on it, 
we wish to separate its role in terms 
of sampling versus evaluating its log posterior, so we will always denote 
the sampling from $\tilde p$, even when this will mean sampling from the 
current estimate of $q$. 

The proposal of this paper is to replace the stochastic 
KL divergence 
minimization with a simpler and 
(as we will show) less noisy ${\rm L_2}$ optimization. 
Since KL divergence enjoys many information theory based 
guarantees, we would also like this ${\rm L_2}$ optimization 
to reduce to KL divergence minimization 
in the high sampling limit, if $\tilde{p}=q$. We will 
show later that this is indeed the case. To be slightly 
more general, we can introduce 
expectation of ${\rm L_n}$ distance (to the power $n$) of log posterior between the two 
distributions,
\begin{equation}
{\rm EL_n}(\tilde{p})= \E_{\tilde p}\left(|\mathcal{L}_q-\mathcal{L}_p|^n \right),
\end{equation}
where $\tilde p$ denotes some approximation to $p$. 
These belong to a larger class of f-divergences, $D_f(p,q)=\E_q f(p/q)$, such that KL($p|q$) is for $f(t)=t\ln t$, while for
${\rm EL_n}(p)$ we have $f(t)=t|\ln t|^n$. 

In this paper we will 
focus on $n=2$. We cannot directly minimize the 
$L_2$ distance because we do not know the 
normalization $\ln p(x)$, 
so instead 
we will minimize it up to the unknown 
normalization, 
\begin{equation}
{\rm EL_2O}(\tilde{p})=\argmin_{\mu,\Sigma,\ln \bar{p}}
\E_{\tilde p}[(\mathcal{L}_q-\tilde{\mathcal{L}}_p-\ln \bar{p})^2 ],
\label{EL2O}
\end{equation}
where $\ln \bar{p}$ 
is an 
approximation to $\ln p(x)$ and is a free parameter to be minimized together with $\mu,\, \Sigma$.
Later we will generalize this to higher order derivatives of $\mathcal{L}_p$, for which 
we do not need to distinguish between $\tilde{\mathcal{L}}_p$ and ${\mathcal{L}}_p$.
The choice of $\tilde{p}$ defines the distance. 
We present first the version where $\tilde p=q$, since we know how to sample from it, later we will generalize it to other sampling proposals. However, 
we view the sampling distribution as unrelated to the 
hyper-parameters of $q$
we optimize for, even when $\tilde{p}=q$,
so unlike ADVI we will not be propagating the gradients with respect to the samples $z_k$
inside $\mathcal{L}_p$. For the 1D Gaussian case
we have
\begin{equation}
{\rm EL_2O}(\tilde{p})=\argmin_{\mu,\Sigma,\ln \bar{p}}N_k^{-1}\sum_k \left[\frac{(z_k-\mu)^2}{2\Sigma}-\tilde{\mathcal{L}}_p(z_k)-c\right]^2. 
\label{elo1}
\end{equation}
where $c=\ln \bar{p}-(\ln 2\pi \Sigma)/2$ is a constant to be optimized together with $\mu$ and $\Sigma$. Note that we are not using equation 
\ref{zk} to simplify the first term: we are separating the role of 
$\tilde{p}$ as a sampling proposal, from $q$ as an approximation to 
$p$, even when $\tilde p=q$. 

\begin{figure}[t!]
\centering \includegraphics[height=0.27\textwidth]{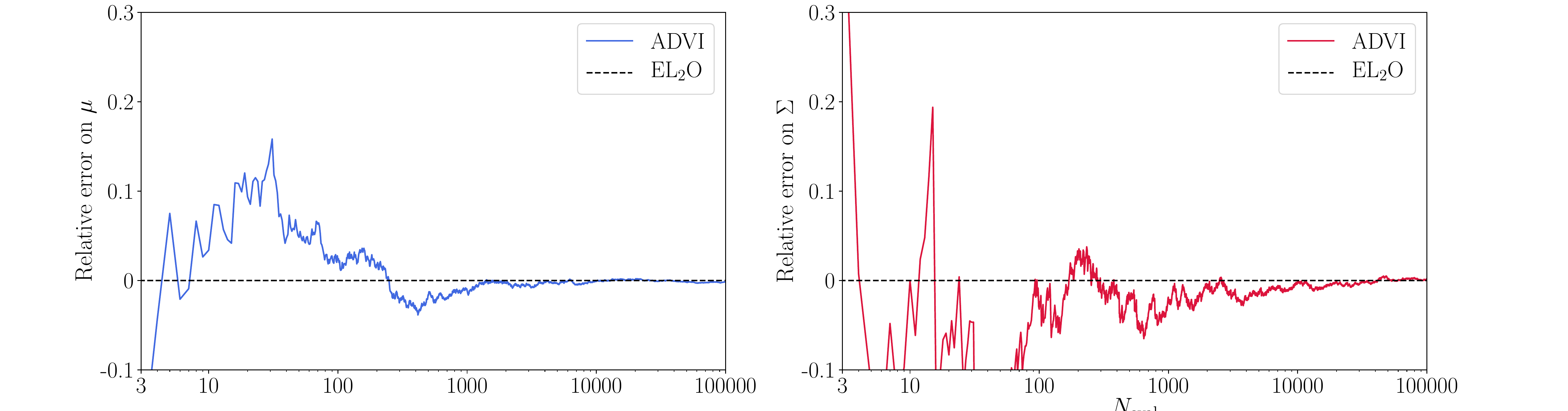}
\caption{Relative errors on the mean $\mu$ and variance $\Sigma$ for the Gaussian ansatz of $q$ in a setting where $p$ is Gaussian. We find that the ADVI solution is noisy and only slowly converges to the correct answer, while EL$_2$O gives the exact solution after 3 evaluations, since there are 3 parameters to determine, after which the problem is over-determined.}
\label{fig:ADVI}
\end{figure}

We see that equation \ref{elo1} is a standard linear regression problem
with polynomial basis up to quadratic order in $z$, and 
the linear parameters to solve are
$c-\mu^2/2\Sigma$ for $z^0$, $\mu/\Sigma$ for $z$
and $-1/2\Sigma$ for $z^2$. If we have $N_k=3$ 
one can obtain the complete solution via normal equations of linear algebra (or a single Newton update if using optimization), and then transform these to determine 
$\mu$, 
$\Sigma$ and $\ln \bar{p}$, which are uniquely determined, and if 
$p$ is Gaussian ($\mathcal{L}_p$ quadratic) ${\rm EL_2O}$ is zero.  
If $N_k>3$ the problem is over-constrained: if $\mathcal{L}_p$ is quadratic in 
$z$
we are not gaining any additional information and ${\rm EL_2O}$ is still zero. 
In fact, in this case the three samples could 
have been drawn from any distribution. There is no sampling noise in minimizing ${\rm EL_2O}$ if $p$ is covered by $q$. 
The results of KL divergence minimization implemented by ADVI and EL$_2$O minimization are shown in figure \ref{fig:ADVI}. 

If $\mathcal{L}_p$ is not quadratic then minimizing ${\rm EL_2O}$ finds the solution that 
depends on the
class of functions $q$ and also on the 
samples drawn from $\tilde{p}$. In this case it is useful that the samples 
are close to the true $p$, since that means that we are weighting ${\rm EL_2O}$ according to 
the true sampling density (we will discuss the optimal choice of $\tilde{p}$ in section \ref{sec4}). 
Even in this situation however minimizing ${\rm EL_2O}$ has 
advantages. We will show below that the optimization is fast, more so 
if $q$ comes close to covering $p$. 
Moreover, the value of ${\rm EL_2O}$ at the minimum is informing us of the quality of 
the fit: if the fit is good and ${\rm EL_2O}$ is low we can be more confident of the resulting $q$ being a good approximation. If the fit is poor we may 
want to look for improvements in $q$. We will discuss several types of these
that can reduce ${\rm EL_2O}$ beyond the full rank Gaussian approximation for $q$.

Even though we focus on ${\rm L_2}$ distance in this paper it is worth 
commenting on other distances. Of particular interest is ${\rm L_1}$
optimization defined as 
\begin{equation}
{\rm EL_1O}(\tilde{p})=\argmin_{\mu,\Sigma,\ln \bar{p}}
\E_{\tilde p}(|\mathcal{L}_q-\tilde{\mathcal{L}}_p-\ln \bar{p}| ).
\end{equation}
If we use $\tilde{p}=q$ then this differs from ${\rm KL}(q||p)$ minimization only 
in taking the absolute value  $|\mathcal{L}_p-\mathcal{L}_q|$
(in terms of f-divergence $t\ln t$ is replaced with $t|\ln t|$).
The difference is that while minimizing ${\rm KL}(q||p)$
minimizes $\mathcal{L}_p$ 
regardless of the $z_k$ dependent part of $\mathcal{L}_q$, ${\rm EL_1O}$ tries to set it to 
$\mathcal{L}_p=\mathcal{L}_q$, up to the normalization $\ln \bar{p}$. 
For a finite number of samples the two solutions differ and the 
latter enforces the sampling variance cancellation, since the 
values of $\mathcal{L}_q$ and $\tilde{\mathcal{L}}_p$ are both evaluated
at the same sample $z_k$. We see from this example that the noise in 
KL divergence can be traced to the fact that its integrand is not required to be 
positive, while it is for ${\rm EL_1O}$ and ${\rm EL_2O}$. While there are 
many f-divergences that have this property, 
${\rm EL_2O}$  leads to 
identical equations as KL divergence minimization 
in the high sampling limit, so for the 
rest of the paper we will focus on ${\rm EL_2O}$ only.

While $t\ln t$ and $t\ln^2t$ f-divergence minimization seems very similar, they are fundamentally different
optimization procedures. Minimization of
KL divergence only makes sense in the context of the KL divergence integral $\int dz q (\ln q -\ln p)$
: it is only positive after the integration and one cannot minimize the integrand
alone. Deterministic integration is only feasible in very low dimensions, and stochastic integration via Monte Carlo converges slowly, as $N_k^{-1/2}$. In contrast, minimizing EL$_2$O is based on comparing $\ln q(z_k)$ and $\ln p(z_k)$ at the same sampling 
points $z_k$: if the two distributions are to be equal they should agree at every sampling point, up to the 
normalization constant. There is no need 
to perform an integral and instead it should be viewed as a loss function
minimization procedure: there is no stochastic integration noise. 


\section{Expectation with $\bf L_2$ optimization ({\bf EL$_2$O}) method}
\label{sec3}

In this section we generalize the EL$_2$O concept in several directions. 
An important trend in the modern statistics and machine learning (ML) applications 
in recent years has been 
the development of automatic differentiation (and Hessians) of the loss function $\mathcal{L}_p$. 
Gradients enable us to do gradient based optimization and sampling, which is the 
basis of recent successes in statistics and ML, from HMC to neural networks. 
Codes such as Tensorflow, PyTorch and Stan have been developed to obtain analytic gradients using backpropagation method. 
The key property of 
these tools is that often the calculational cost of the analytic gradient 
is comparable to the cost of evaluating the function itself: this is because the
calculation of the function and its gradient share many components, such that 
the additional operations of the gradient do not significantly increase the overall 
cost. In some cases, such as the nonlinear least squares to be discussed below, 
a good approximation of the 
Hessian can be obtained using function gradients alone (Gauss-Newton approximation), 
so if one has the function gradient one also gets an approximation to the Hessian for free. 
When one has access to the 
gradients and Hessian it is worth taking advantage of this information for 
posterior inference.  

Assume we have a general gradient expansion of log posterior around a
sample $\bi{z}_k$, 
\begin{equation}
\mathcal{L}_p(\bi{z}_k+\Delta\bi{z}_k)= 
\sum_{n=0}^{\infty} \frac{1}{n!}\nabla^n_{\bi{z}}\mathcal{L}_p(\bi{z}_k)(\Delta\bi{z}_k)^n, 
\end{equation}
where $\nabla^n_{\bi{z}}\mathcal{L}$ is $M^n$ dimensional tensor of higher order derivatives. 
For $n=0$ this the log posterior value, for $n=1$ this is its gradient vector and for $n=2$ this is its Hessian matrix. 

We perform the same expansion for our approximate posterior $q(\bi{z},\bi{\theta})$, 
\begin{equation}
\mathcal{L}_q
(\bi{z}_k+\Delta\bi{z})= 
-\ln q(\bi{z}_k+\Delta\bi{z})= \sum_{n=0}^{\infty} 
\frac{1}{n!}\nabla^n_{\bi{z}}\mathcal{L}_q(\bi{z}_k)(\Delta\bi{z})^n.
\end{equation}
We assume that $q(\bi{z},\bi{\theta})$ has a simple form so that these gradients can be 
evaluated analytically, and that it depends on hyper-parameters $\bi{\theta}$ we wish to 
determine. 
We will begin with 
a multivariate Gaussian assumption for $q(\bi{z})$
 with mean $\bi{\mu}$ and covariance $\bi{\Sigma}$,
\begin{equation}
q(\bi{z})=N(\bi{z};\bi{\mu},\bi{\Sigma})=(2\pi)^{-N/2}\det \bi{\Sigma}^{-1/2}e^{-\frac{1}{2}(\bi{z}-\bi{\mu})^T\bi{\Sigma}^{-1}(\bi{z}-\bi{\mu})},
\label{fullrank}
\end{equation}
\begin{equation}
\mathcal{L}_q =\frac{1}{2}\left[\ln \det \bi{\Sigma}+(\bi{z}-\bi{\mu})^T\bi{\Sigma}^{-1}(\bi{z}-\bi{\mu})+N\ln(2\pi) \right].
 \label{lnq}
 \end{equation}
To make the expansion coefficients dimensionless we
can first scale  $z_i$, 
\begin{equation}
   z_i \rightarrow \frac{z_i-\mu_i}{\Sigma_{ii}^{1/2}}.
   \label{scale}
\end{equation}
While centering (subtracting $\mu_i$)  is not required at this stage, 
we will use it when we discuss non-Gaussian posteriors later. 
This scaling can be done using some approximate $\Sigma_{ii}$ from $q$
that need not be iterated upon, so for this reason 
we will in general separate it from 
the iterative method of determining $q$. In practice we 
set the scaling $\Sigma_{ii}$ after the burn-in 
phase of iterations, when $q$ is
typically determined by MAP and 
Laplace approximation. 

The considerations in previous section, combined 
with the availability of analytic 
gradient expansion terms, suggest to generalize  ${\rm EL_2O}$ to the form that is the 
foundation of this paper,
\begin{equation}
{\rm EL_2O}=\argmin_{\bi{ \theta}} \E_{\tilde p} \left\{N_{\rm der}^{-1}\sum_{n=0}^{n_{\rm max}} \sum_{i_1,..i_n}\alpha_n\left[
\nabla^n_{\bi{z}}\mathcal{L}_q(\bi{z})-\nabla^n_{\bi{z}}\mathcal{L}_p(\bi{z},\bi{\theta})
\right]^2\right\},
\label{argmin}
\end{equation}
where the averaging is done over the samples $\bi{z}_k$ drawn from $\tilde p$, and 
where we define, for $n=0$, $\mathcal{L}_p=\tilde{\mathcal{L}}_p+\ln \bar{p}$. The sum over 
index $i_1..i_n$ should be symmetrized, so for example for $n=2$ it is over 
$i_1, i_2\ge i_1$. Here $N_{\rm der}$ is the total 
number of all the terms, while $n_{\rm max}$ is the largest order of the 
derivatives we wish to include. 
This
equation combines three main elements: sampling from
$\tilde p$, evaluation of the log posterior $\mathcal{L}_p$
and its derivatives at samples, analytic evaluation of the same for $\mathcal{L}_q$,
and finally $L_2$ objective optimization to find the 
best fit parameters $\bi{\theta}$. The latter involves evaluating 
another sequence of first (and second, if second order optimization is used) order gradients, this time with respect to $\bi{\theta}$. 

In equation \ref{argmin} we have introduced the
weight $\alpha_n$, which accounts for different weighting of different derivative order, so that for example 
each tensor element of a Hessian $\nabla_{z_i}\nabla_{z_j}\mathcal{L}_p$ can have a different weight to each 
vector element of a gradient $\nabla_{z_i}\mathcal{L}_p$, which can have different weight as each $\mathcal{L}_p$. 
Because of scaling in equation \ref{scale} the
expression is invariant under reparametrization of $\bi{z}$, and one can view each element of gradient and 
Hessian as one additional evaluation of $\mathcal{L}_p$, which 
should have equal weight. For example, one can view $\mathcal{L}_p$ and $\nabla_{z_i}\mathcal{L}_p$
using a finite difference expression of the gradient into an evaluation of two $\mathcal{L}_p$
at two independent samples. In this view equation \ref{EL2O} turns into equation \ref{argmin}. However, 
because we want the samples to be spread over sampling proposal $\tilde{p}$, it is suboptimal to have $N$ "samples" at 
nearly the 
same point gives by the gradient information, and for this reason $\alpha_0>\alpha_1>\alpha_2$. 

Equal weight can also be justified using the fact that different gradient order terms determine 
different components of $\bi{\theta}$. 
In the 
context of a full rank Gaussian for $q$ we 
can think of the log likelihood determining an 
approximation to the normalization constant $\ln p(x)$, 
the gradient determining the means $\bi{\mu}$, and the 
Hessian determining the covariance $\bi{\Sigma}$, of 
which the 
latter two
are needed for the posterior. We will in fact write the 
optimization equations below explicitly in the form where these terms are 
separated. 
Due to the sample variance cancellation 
a single sample evaluation suffices, but 
each additional evaluation of these variables
can be used to improve the proposed $q$. For example, in a Gaussian mixture model for 
$q$, at each sample evaluation of the log posterior, its gradient and Hessian gives enough 
information to fit another full rank Gaussian mixture component. 
However, other 
weights may be worth exploring, such as downweighting the Hessian 
in situations where it is only 
approximate, as in the Gauss-Newton approximation discussed further below. 

We will generally stop at $n_{\rm max}=2$, but 
in some circumstances 
 having access
to analytic gradients beyond the Hessian could be beneficial. We will see below
that one of the main problems of full rank Gaussian variational methods is its inability to 
model the change of sign of Hessian off-diagonal elements, which can be 
described with the third order expansion ($n_{\rm max}=3$) terms. 
However, 
doing gradient expansion around a single point has its limitations: the shape of 
the posterior could be very different just a short distance away:
there is no substitute for 
sampling over the entire probability distribution. Moreover, evaluations beyond the 
Hessian may be costly even if analytic derivatives are used. 
For this reason 
we will develop here three different versions depending on whether we have 
access to only $n_{\rm max}=0$ information, $n_{\rm max}=1$, or $n_{\rm max}=2$. 

\subsection{Gradient and Hessian version}

For $n_{\rm max}=2$ we optimize 
\begin{equation}
{\rm EL_2O}=N_M^{-1}\argmin_{\ln \bar{p},\bi{ \mu},\bi{\Sigma}^{-1}}\E_{\tilde p}\left\{ \sum_{i,j\le i}^M\left[\nabla_{z_i}\nabla_{z_j}\mathcal{L}_q-\nabla_{z_i}\nabla_{z_j}\mathcal{L}_p\right]^2+\sum_{i=1}^M \left[\nabla_{z_i}\mathcal{L}_q-\nabla_{z_i}\mathcal{L}_p\right]^2
+[\mathcal{L}_q-\tilde{\mathcal{L}}_p-\ln \bar{p}]^2
\right\},
\label{argmin1}
\end{equation}
where $N_M=M(M+3)/2+1$.
In this section 
we will drop for simplicity the last term
and optimization 
over $\ln \bar{p}$ since 
we do not need it
($q$ is already normalized). For a more 
general $q$ such as a Gaussian mixture model this term needs to be included, as discussed 
further below. 
There is additional flexibility in terms of how much weight to give to Hessian versus 
gradient information, which we will not explore in this 
paper. 

We first need analytic gradient and Hessian information of $q(\bi{z})$. Taking the gradient of $\mathcal{L}_q$ in
equation \ref{lnq} with respect to $\bi{z}$ gives
\begin{equation}
\nabla_{\bi{z}}\mathcal{L}_q=\bi{\Sigma}^{-1}(\bi{z}-\bi{\mu}).
\label{qgradz}
\end{equation}
The Hessian is obtained as a second derivative of $\mathcal{L}_q(\bi{z})$ with respect to $\bi{z}$, 
\begin{equation}
\nabla_{\bi{z}}\nabla_{\bi{z}}\mathcal{L}_q(\bi{z})=\bi{\Sigma}^{-1}.
\label{qgrad2z}
\end{equation}

Even if we have only a single sample we can expect the Hessian evaluated at the 
sample to determine $\bi{\Sigma}$
(since it has no dependence on $\bi{\mu}$), and with $\bi{\Sigma}$ determined we can use
its gradient to determine $\bi{\mu}$ from equation \ref{qgradz}. In this approach we can thus write
the optimization solution of equation \ref{argmin1} as 
\begin{equation}
\bi{\Sigma}^{-1}=\E_{\tilde p} \left[ \nabla_{\bi{z}} \nabla_{\bi{z}} \mathcal{L}_p  \right]
\approx
N_{k}^{-1} \sum_{k=1}^{\rm N_k}\nabla_{{\bi z}} \nabla_{\bi{z}} \mathcal{L}_p (\bi{z}_{k}). 
\label{Laplace}
\end{equation}

Applying the optimization of equation \ref{argmin} with respect to $\bi{\mu}$ and
keeping the gradient terms only (i.e. dropping $n=0$ term, since $n=2$ term has no $\bi{\mu}$
dependence) we find, 
\begin{equation}
\E_{\tilde p}[\nabla_{\bi{z}}\mathcal{L}_p(\bi{z})]=\bi{\Sigma}^{-1}(\E_{\tilde p} [\bi{z}] -\bi{\mu}), 
\end{equation}
\begin{equation}
\bi{\mu}=-\bi{\Sigma}\E_{\tilde p} [ \nabla_{\bi{z}}\mathcal{L}_p(\bi{z})+\bi{z}]\approx 
N_{\rm k}^{-1}\sum_{k=1}^{\rm N_k}\left[-\bi{\Sigma}\nabla_{\bi{z}} \mathcal{L}_p (\bi{z}_{k})+\bi{z}_{k}\right]. 
\label{mu2}
\end{equation}

Expectation of these equations has been derived in the context of variational methods
\citep{OpperArchambeau09}, showing that the solution to ${\rm EL_2O}$ is the same as VI minimization 
of ${\rm KL}(q||p)$ in the high sample limit, if $\tilde p=q$. But there is a difference in the 
sampling noise if the number of samples is low: the presence of $\bi{z}_{k}$ at the 
end of equation \ref{mu2} guarantees there is no sampling noise, and no such term
appears in stochastic KL divergence based minimization. 
As we argued above, stochastic minimization of KL divergence has sampling 
noise, while minimizing ${\rm EL_2O}$
gives estimators in equations \ref{Laplace}, \ref{mu2} that 
are exact even for a single sample, under the 
assumption of the posterior belonging to the family of model posteriors 
$q(\bi{z})$: it does not even matter where we draw the sample. 
If the posterior 
does not belong to this family we need to perform the expectation in equations \ref{Laplace}, \ref{mu2},
by averaging over more than one sample. There will be sampling noise, but the closer $q$ family is to the posterior $p$
the lower the noise. 
This will be shown explicitly in examples of section \ref{sec4}, 
where we test the method on $q$'s generalized beyond the full 
rank Gaussian. 
The residual ${\rm EL_2O}$ informs us when this is needed: if it 
is large it
indicates the need to improve $q$, by going beyond the full rank Gaussian.
The approach of this paper is to generalize $q$ until we 
find a solution with low residual
${\rm EL_2O}$ so that the posterior is reliable: in examples of section \ref{sec4} 
this is reached approximately when ${\rm EL_2O}< 0.2$. 

Since the gradient and Hessian gives enough information to determine $\bi{\mu}$ and 
$\bi{\Sigma}$ we can convert this into an iterative process where we draw a few 
samples, even as low as a single sample only.
Assume that at the current iteration the estimate is $\bi{\mu}_t$ and 
$\bi{\Sigma}_t$
and that we have drawn a single sample $\bi{z}_1$ from $q=N(\bi{z};\bi{\mu}_t,\bi{\Sigma}_t)$.
We also evaluate the gradient $\nabla_{\bi{z}}\mathcal{L}(\bi{z}_1)$ 
and Hessian, giving the following updates 
\begin{equation}
\bi{\Sigma}^{-1}_{t+1}=\nabla_{\bi{z}}\nabla_{\bi{z}}\mathcal{L}(\bi{z}_1),
\end{equation}
\begin{equation}
\bi{\mu}_{t+1}=-\bi{\Sigma}_{t+1}\nabla_{\bi{z}}\mathcal{L}(\bi{z}_1)+\bi{z}_1. 
\end{equation}
With an update $\bi{\mu}_{t+1}$ we can draw a new sample and repeat the process
until convergence. In this paper we are assuming 
that the Hessian inversion to get the covariance 
matrix and sampling from it via Cholesky decomposition 
is not a computational bottleneck. This would limit the method 
to thousands of dimensions if full rank description of all 
the variables is needed, and if the cost 
of evaluating $\mathcal{L}_p$ is moderate. 
Note that if we were doing MAP we would have assumed $q$ is a delta function with 
mean $\bi{\mu}_t$, in which case there is only one sample at $\bi{z}_1=\bi{\mu}_t$: 
the equation above becomes equivalent to a 
second order Newton's method for MAP optimization. One obtains a full distribution in the 
full rank Gaussian approximation
by evaluating the Hessian at the MAP estimate (Laplace approximation). 
The ${\rm EL_2O}$ 
method is a very simple generalization of the Laplace approximation 
and for a single sample 
has equal cost, as long as 
the cost of matrix inversion is low. 
Once we have approximately converged on $\bi{\mu}$ (burn-in phase) we can average over more 
samples to obtain a more reliable approximation for both $\bi{\mu}$ and 
$\bi{\Sigma}$. This often gives a more reliable estimate of $\bi{\mu}$ since 
it smooths out any small scale ruggedness in the posterior. Typically we find a few samples suffice
for simple problems.
When problems are not simple (and ${\rm EL_2O}$ remains high) 
it is better to increase the 
expressiveness of $q$ beyond the full rank Gaussian, as discussed below.
Only the full rank Gaussian
allows analytic marginalization of correlated variables: 
one inverts the Hessian matrix $\bi{\Sigma}^{-1}$ to obtain the 
covariance matrix $\bi{\Sigma}$, and marginalization is 
simply eliminatation of the rows 
and columns of $\bi{\Sigma}$ 
for the parameters we want to marginalize over (for proper normalization one also needs to 
evaluate the determinant of the remaining sub-matrix). 
We want to 
preserve this property for more general $q$ as well, and given the above 
stated property of full rank Gaussian two ways to do so are one-dimensional 
transforms and Gaussian mixtures, both of which will be discussed below. 

\begin{algorithm}[tb]
   \caption{Full rank Gradient and Hessian  EL$_2$O}
   \label{alg1}
\begin{algorithmic}
   \STATE {\bfseries Input:} data $x_{i}$, size $N$
   \STATE Initialize parameters $z_i$: random sample from prior, size $M$
   \STATE Find MAP using optimization for initial $\bi{\mu}$. Use Laplace for initial $q(\bi{z})=N(\bi{\mu},\bi{\Sigma})$.
   \WHILE{EL$_2$O value has not converged}
   \STATE Draw a new sample $z_{N_k+1}$. Increase $N_k$ by 1.  
   \IF{Hessian available}
   \STATE $\bi{\Sigma}^{-1}=N_{k}^{-1} \sum_{k=1}^{\rm N_k}\nabla_{{\bi z}} \nabla_{\bi{z}} \mathcal{L}_p (\bi{z}_{k})$
   \ELSE{}
   \STATE $\bi{\mathcal{H}}=\sum_{k=1}^{\rm N_k}(\bi{z}_k-\bi{\mu})(\bi{z}_k-\bi{\mu})$
   \STATE $\bi{\Sigma}^{-1}= \bi{\mathcal{H}}^{-1}\sum_{k=1}^{\rm N_k} (\bi{z}_k-\bi{\mu})\nabla_{\bi{z}}\mathcal{L}(\bi{z}_k) $
   \ENDIF
   \STATE $\bi{\mu}=N_{\rm k}^{-1}\sum_{k=1}^{\rm N_k}\left[-\bi{\Sigma}\nabla_{\bi{z}} 
   \mathcal{L}_p (\bi{z}_{k})+\bi{z}_{k}\right]$
   \STATE Compute EL$_2$O
   \ENDWHILE
\end{algorithmic}
\end{algorithm}

\subsection{Gradient only and gradient free versions}

We argued above that it is always beneficial to evaluate the Hessian, since together 
with the gradient this gives us 
an immediate estimate of $M(M+3)/2$ parameters, which can be chosen to be 
$\bi{\Sigma}^{-1}$ and $\bi{\mu}$, and so we get a full rank $q$
with a single sample. 
Moreover, 
for nonlinear least squares and related problems evaluating the Hessian in Gauss-Newton 
approximation is no more expensive than evaluating the gradient. Suppose however that 
the Hessian is not available and we only have access to the gradients.
This is for example the ADVI strategy \citep{KucukelbirTRGB17}, but let us look 
at what our approach gives. Specifically, we want to minimize
\begin{equation}
{\rm EL_2O}=(M+1)^{-1}\argmin_{\ln \bar{p},\bi{ \mu},\bi{\Sigma}^{-1}} \E_{\tilde p}\left[ 
\sum_{i=1}^M \left\{\nabla_{z_i}\mathcal{L}_q-\nabla_{z_i}\mathcal{L}_p\right\}^2+
 \left\{\mathcal{L}_q-\tilde{\mathcal{L}}_p-\ln \bar{p}\right\}^2
\right],
\label{grad}
\end{equation}
where again for simplicity of this section we will drop the last term and not optimize 
over $\ln p(\bi{x})$, since we do not need 
it. The first term on the RHS is called the Fisher divergence $F(q,p)$ if sampled
from $q$ and $F(p,q)$ if sampled from $p$ \citep{Hammad78}, and Jensen-Fisher 
divergence if averaged over the two \citep{SanchezZarzo12}. 
This shows the connection of the gradient 
part of ${\rm EL_2O}$ to the Fisher information. 

First derivatives with respect to $\bi{\mu}$ give equation \ref{mu2}. To evaluate it 
we need to determine $\bi{\Sigma}$. 
To get the equation for $\bi{\Sigma}$ we first derive its gradient 
$ \nabla_{\bi{\Sigma}^{-1}}{\rm EL_2O}$,  and its
Hessian, $\bi{\mathcal{H}}=\nabla_{\bi{\Sigma}^{-1}} \nabla_{\bi{\Sigma^{-1}}}{\rm EL_2O}$,  
\begin{equation}
\bi{\mathcal{H}}=\E_{\tilde p} [ (\bi{z}-\bi{\mu})(\bi{z}-\bi{\mu})]
\approx \sum_{k=1}^{\rm N_k}(\bi{z}_k-\bi{\mu})(\bi{z}_k-\bi{\mu}) . 
\end{equation}
We need $N_k=M+1$ gradients sampled at $\bi{z}_k$ for this matrix to be non-singular if $\bi{\mu}$
is also determined from the same samples. Taking the 
first derivative of equation \ref{grad} with respect to $\bi{\Sigma}^{-1}$ gives
\begin{equation}
\bi{\Sigma}^{-1}=\bi{\mathcal{H}}^{-1}
\E_{\tilde p} \left[
(\bi{z}-\bi{\mu})\nabla_{\bi{z}}\mathcal{L} \right]
\approx \bi{\mathcal{H}}^{-1}\sum_{k=1}^{\rm N_k} (\bi{z}_k-\bi{\mu})\nabla_{\bi{z}}\mathcal{L}(\bi{z}_k) .
\label{gradsig}
\end{equation}
We obtained a set of equations \ref{mu2} and \ref{gradsig} that only use gradient information, but these are 
different from the ADVI equations \citep{KucukelbirTRGB17}. In particular, our equations 
have sampling variance cancellation built in, and if $q$ covers $p$ they 
give zero error once we have drawn 
enough samples so that the system is not under-constrained.  
For $\tilde{p}=p$
the $L_2$ norm of equation \ref{grad} has also been proposed by \cite{Hyvarinen05} as a 
score matching statistic, but was rewritten 
through integration by parts into a form that does not cancel sampling variance, 
similar to ${\rm KL}(p||q)$.

Finally, if we have no access to gradients we can still apply the ${\rm EL_2O}$ method: 
to get all the 
full rank parameters we need to evaluate the loss function in $M(M+3)/2+1$ points, and then 
optimize
\begin{equation}
{\rm EL_2O}=\argmin_{\ln \bar{p},\bi{ \mu},\bi{\Sigma}^{-1}}\E_{\tilde p}\left[ \left\{\tilde{\mathcal{L}}_q-\mathcal{L}_p-\ln \bar{p}\right\}^2\right],
\label{gradfree}
\end{equation}
where this time we also need to optimize for $\ln \bar{p}$ 
together with $\bi{\mu}$ and $\bi{\Sigma}$. 
These equations also incorporate the sampling variance cancellation.

Hybrid approaches are also possible: for example, we may have access to analytic first or 
second derivatives 
for some parameters, but not for others. In this case, one can design an optimization 
process that uses analytic gradients and Hessian components for some of the parameters, while relying on 
either numerical finite differences or gradient free approaches
for the other parameters. More generally, some parameters may require 
expensive and slow evaluations (slow parameters) while others can be inexpensive (fast parameters).
In this case, we can afford to do numerical gradients with respect to fast parameters and 
focus on development of analytic gradients for slow parameters. Another hybrid
approach will be discussed in the context of Gauss-Newton approximation below, 
where we use the Hessian in the Gauss-Newton approximation for the covariance 
matrix $\bi{\Sigma}$, and only the gradient for the remaining hyper-parameters of $q$. 

\subsection{Posterior expansion beyond the full rank Gaussian: bijective 1D transforms}

So far we obtained the full rank VI solution with an iterative process which should 
converge nearly as rapidly as MAP. If the posterior is close to the assumed multi-variate 
Gaussian then this process converges fast, and only a few samples are needed. If there is 
strong variation between the Hessian elements evaluated at different sampling points then we know
the posterior is not well described by a multi-variate Gaussian. In this case we may 
want to consider proposal functions beyond equation \ref{fullrank}. 
However, a multi-variate Gaussian is the only 
correlated multi-variate distribution where analytic marginalization can be done by simply 
inverting the Hessian matrix. This is a property that we do not want to abandon. 
For this reason we will first consider one-dimensional transformations of the original 
variables $\bi{z}$ in this subsection, and Gaussian mixtures in the next. 
Variable transformations need to 
be bijective so that we can easily go from one set of the variables to the other and back \citep{RezendeMohamed15}. 
Here we will 
use a very simple family of models  that give rise to 
skewness and curtosis, which are the one-dimensional versions of the gradient 
expansion at third and fourth order. 

Specifically, we will consider bijective transformations of the form $y_i(z_i)$ such that 
\begin{equation}
q(\bi{z})=N({\bi{y}};\bi{\mu}, \bi{\Sigma})\Pi_i |J_i|,\,\,J_i=\frac{dy_i}{dz_i}, 
\label{multiy}
\end{equation}
with $N({\bi{y}};\bi{\mu}, \bi{\Sigma})$ given by equation \ref{fullrank}. Under this form the marginalization 
over the variables is trivial. For example, 
marginalized posterior distribution of $z_i$ is \newline $q[z_i(y_i)]=N({y_i};\mu_i,\Sigma_{ii})|dy_i/dz_i|$, 
where $\Sigma_{ii}$ is the diagonal component of the covariance matrix, obtained by 
inverting the Hessian matrix $\bi{\Sigma}^{-1}$. 

We would like to modify the variables $z_i$ such that 
the resulting posteriors can accommodate more of the variation of $\tilde{\mathcal{L}}_p$. 
In one dimension this would be their skewness and curtosis, which are indicated for 
example by the variation of the Hessian with the sample, but in higher dimensions 
we also want to accommodate variation of off-diagonal terms of the Hessian. 
In typical situations given the full rank 
solution and the scaling of equation \ref{scale} 
the posterior mass 
will be concentrated around $-1<z_i<1$, but the distribution may be skewed, or have 
more or less posterior mass outside this interval. 

A very simple change of variable is 
$
y_i=z_i+\frac{1}{2}\epsilon_i z_i^2+\frac{1}{6}\eta_i z_i^3$,
where we assume $\epsilon_i$ and $\eta_i$ are both parameters that can be either 
positive or negative, but small such that the relation is invertible. In one dimension 
the log of
posterior is, keeping the terms at the lowest order in $\epsilon_i$ and $\eta_i$, 
$2\mathcal{L}_q \approx c+z_i^2+\epsilon_i z_i^3+\frac{1}{3}\eta_i z_i^4...$
Viewed as a Taylor expansion 
we see that $\epsilon_i$ term determines the third order gradient expansion 
and $\eta_i$ the fourth order, both around $z_i=0$. 

To make these expressions valid for larger values of $\epsilon_i$ and $\eta_i$ we promote 
the transformation into 
\begin{equation}
y_i(z_i)={\rm sinh}_{\eta}\left[ \frac{\exp(\epsilon_iz_i)-1}{\epsilon_i}\right],
\label{sk}
\end{equation}
where for $\epsilon_i=0$ the above is just $y_i(z_i)={\rm sinh}_{\eta} z_i$ \citep{SchuhmannJoachimiEtAl16}
\begin{equation}
{\rm sinh}_{\eta}(x) =
\begin{cases}
\eta^{-1}{\rm sinh}(\eta x) & (\eta>0) \\ x & (\eta=0) \\ \eta^{-1}{\rm arcsinh}(\eta x) & (\eta<0).
\end{cases}
\end{equation}
These are bijective, but not guaranteed to give the required posteriors.
We can however apply the transformations multiple times for a more expressive 
family of models. For small values of 
$\epsilon$ and $\eta$ equation \ref{sk} reduces to the Taylor expansion above.
If the posteriors are multi-peaked then these transformations may not be sufficient, 
and 
Gaussian mixture models can be used instead, discussed below. 

The gradient of $\mathcal{L}_q $ is 
\begin{equation}
\nabla_{z_i}\mathcal{L}_q=\sum_j(\Sigma^{-1})_{ij}(y_j-\mu_j) J_i-\frac{\nabla_{z_i}|J_i|}{|J_i|}, 
\end{equation}
while the Hessian is 
\begin{equation}
\nabla_{z_i}\nabla_{z_j}\mathcal{L}_q=(\Sigma^{-1})_{ij}J_iJ_j+\left[\sum_k (\Sigma^{-1})_{ik}(y_k-\mu_k) \nabla_{z_i}J_i -\frac{\nabla_{z_i}\nabla_{z_i}|J_i|}{|J_i|}+\left(\frac{\nabla_{z_i}|J_i|}{|J_i|}\right)^2\right]\delta_{ij}. 
\label{hnl}
\end{equation}

If the Hessian is varying with the samples $\bi{z}_k$ we have an indication 
that we need higher order corrections. 
With the gradient and Hessian at a single sample $\bi{z}_1$ we 
have the sufficient number of constraints, $M(M+3)/2$, to determine $\bi{\mu}$ and $\bi{\Sigma}^{-1}$. If we evaluate these variables at another sample $\bi{z}_2$ we 
already have too many constraints to determine the additional $2M$ nonlinear transform variables, so the problem 
is overconstrained even with two drawn samples. This is the power of having access to the gradient and 
Hessian information: we converge fast both because we can use Newton's
method to find the solutions and because a few samples give us enough 
constraints. 

\subsection{Posterior expansion beyond the full rank Gaussian: Gaussian mixtures}
A second non-bijective way that can extend the expressivity of posteriors while still allowing for analytic 
marginalizations is a Gaussian mixture model \citep{BishopLawrenceEtAl97}. Here we model the posterior as a weighted sum 
of several multi-variate Gaussians, each of which can have an additional 1D NL transform, as 
in equation \ref{multiy}, 
\begin{equation}
q(\bi{z})=\sum_j w_j N({\bi{y}^j};\bi{\mu}^j, \bi{\Sigma}^j)\Pi_i \left|\frac{dy_i^j}{dz_i}\right|
\equiv \sum_j w_j q^j(\bi{z}),
\label{gm}
\end{equation}
where $\sum_j w_j=1$. 
We can introduce the position dependent weights
\begin{equation}
w_j(\bi{z})=\frac{w_jq^j(\bi{z})}{q(\bi{z})}.
\end{equation}

We can now derive the corresponding $\nabla_{\bi{z}}\mathcal{L}_q$ and $\nabla_{\bi{z}}\nabla_{\bi{z}}\mathcal{L}_q$. For example, if $\bi{y}=\bi{z}$ the gradient is 
\begin{equation}
\nabla_{\bi{z}}\mathcal{L}_q=\sum_j w_j(\bi{z})\nabla_{\bi{z}} \mathcal{L}_q=\sum_j w_j(\bi{z})(\bi{\Sigma}^j)^{-1}(\bi{z}-\bi{\mu}^j), 
\end{equation}
and is simply a weighted gradient of each of the Gaussian mixture components. The Hessian is
\begin{eqnarray}
&&\nabla_{\bi{z}}\nabla_{\bi{z}}\mathcal{L}_q=\sum_j \left[\nabla_{\bi{z}}w_j(\bi{z})\nabla_{\bi{z}} 
\mathcal{L}_q+w_j(\bi{z})\nabla_{\bi{z}}\nabla_{\bi{z}}\mathcal{L}_q\right]\nonumber \\
&=&\sum_j w_j(\bi{z})
(\bi{\Sigma}^j)^{-1}-\sum_i\sum_{j \ne i} \frac{w_i(\bi{z})w_j(\bi{z})}{w_i}
\left[(\bi{\Sigma}^j)^{-1}(\bi{z}-\bi{\mu}^j)(\bi{\Sigma}^i)^{-1}(\bi{z}-\bi{\mu}^i)\right].
\label{hessgm}
\end{eqnarray}

Once we have these
expressions we can insert them into equation \ref{argmin} and optimize against the parameters of $q(\bi{z})$. 
This is an optimization problem and requires iterative method to find the solution, but 
no additional evaluations of $\mathcal{L}_p$. While previously we did not use 
$\mathcal{L}_p$ itself since we did not need $c$, now we need to use this value as 
well, at it determines the weights $w_j$. 
With $\mathcal{L}_p$, its gradient and its 
Hessian we have enough data to determine one Gaussian mixture component per sample 
$\bi{z}_k$. 
Once we have constructed the full $q$, 
to analytically marginalize over some of $\bi{z}$ we need to invert separately each of the matrices $(\bi{\Sigma}^j)^{-1}$. 

To summarize, there exist expressive posterior parametrizations beyond the
full rank Gaussian that allow for 
analytic marginalizations and that can fit a broad range of posteriors.
Gaussian mixture model can for example be used for a multi-modal posterior distribution, and a 
single evaluation with a Hessian can fit one component of the multi-variate Gaussian mixture model.
One dimensional nonlinear transforms can be used to give skewness, curtosis and even multi-modality
to each dimension. 

\subsection{Sampling proposals}
\label{sampl}

We argued above that KL$(p||q)$ minimization leads to standard MC method, while 
minimizing KL$(q||p)$ gives VI method. ${\rm EL_2O}$ has more flexibility in terms of the choice of the sampling proposal $\tilde{p}$. We list some of these below, with some 
specific examples presented in section \ref{sec4}. 

{\it Sampling from $q$}: If we sample from $\tilde{p}=q$ and minimize
${\rm EL_2O}$ we get results equivalent to VI in the large sample limit.
While sampling from $q$, and iterating on it, is the simplest choice,
it is not the only choice, and may not be the best choice either. 
One disadvantage is that the samples change as we vary $q$ during optimization, increasing the 
number of calls to $\mathcal{L}_p$. A second disadvantage is that in high 
dimensions sampling from the full rank Gaussian becomes impossible since the 
cost of Cholesky decomposition becomes prohibitive. We will address this 
problem elsewhere. 

{\it Sampling from $p$}: one
alternative is to sample 
from $p$ itself. This has the advantage that the samples do not need to change 
as we iterate on $q$, and if the cost of $\mathcal{L}_p$ is dominant this 
can be an attractive possibility. There are inference problems where sampling from $p$ is 
easy. An example are forward inference problems: suppose we know 
the prior distribution of $\bi{x}$ and we would like to 
know the posterior of $\bi{z}=f(\bi{x})+\bi{n}$, where $f$ is some 
function and $\bi{n}$ is noise. In this case we can create samples of $\bi{z}$ drawn from $p$ simply 
by drawing samples of $\bi{x}$ from its prior and $\bi{n}$ from noise distribution
and evaluating $\bi{z}=f(\bi{x})+\bi{n}$. 

In most cases however sampling from $p$ is hard. 
One possibility is to create a number of true samples from $p$ using
MCMC. This may be expensive, since there is a burn-in period that one 
needs to overcome first. We can use ${\rm EL_2O}$ 
optimization with $\tilde{p}=q$
for the burn-in phase to get to the minimum, and from there sampling can be 
almost immediate, but samples will be correlated, and often 
the correlation length can be hundreds or more. For modern methods like 
HMC sampling can be more efficient in traversing the 
posterior with a lower number of $\tilde{\mathcal{L}}_p$ 
calls, so this approach is worth exploring further. 

{\it Sampling from approximate $p$}: since MCMC is expensive
one can try cheaper alternatives. One possibility is to sample from an approximate posterior generated with the help of 
simulations. Suppose one generates a simulation where one knows the answer. 
One then performs the analysis as on the data, obtaining the point estimate 
on the parameters in terms of their best fit mean 
or mode (MAP). Since we know the truth for simulation we can create a data sample 
by adding the difference between the truth and the point estimate of the 
simulation to the point estimate of the data. This will give an approximation 
to the posterior distribution $p$ where each sample is completely independent. 
It will not be exact because of realization dependence of the posterior, 
and some of the samples may end up being very unlikely in the sense of 
having a very high $\tilde{\mathcal{L}}_p$. Additional importance sampling 
may be needed to improve this posterior further. 

Another strategy is not to sample at all, but devise a deterministic algorithm to select the points where to evaluate 
$\mathcal{L}_p$. 
For example, 
given a MAP+Laplace solution one could select the points that 
are exactly a fixed fraction of standard deviation away from MAP for every parameter. 
This strategy has had some success in filtering applications, where it is called 
unscented Kalman filter (UKF, \cite{JulierUhlman04}). Another option is 
deterministic quadratures, such as Gauss-Hermite integration. These 
deterministic approaches become very expensive in high dimensions. 

{\it Sampling from $q+p$}: symmetric KL divergence is called Jensen-Shannon divergence, and we can similarly do the same for 
EL$_2$O. Since L$_2$ norm is already symmetric we just need 
to sample from $p+q$. 
This may have some benefits: if one samples from $p$ then 
there is no penalty for posterior densities $q$ which do not vanish 
where $p=0$. Conversely, if one samples from $q$ there is no penalty 
for situations where $p$ does not vanish while $q=0$. 
For example, there may be 
multiple posterior peaks in true $p$, but if we only found one we would never know 
the existence of the others. In both cases 
the difficulty arises because the normalization of $p$ is not accessible to us.
The latter case is often argued to be more 
problematic suggesting sampling from $p$ should be used if possible. 
However, note that in the case of widely separated 
posterior peaks sampling from true $p$ using MCMC may not be possible either, as MCMC
may not find all of the posterior peaks. In this case 
sampling from $q$ using 
multiple starting points with 
a Gaussian mixture for $q$ is a
better alternative. 

If both of these issues are a concern one can try to sample from both
$q$ and from $p$, mixing the two types of samples. 
This will give us 
samples where $p=0$ but $q>0$, and where $q=0$ but $p>0$, if 
such regions exist. 
Another
hybrid sampling $p+q$ method is, after the burn in, to iterate on $q$,
sample from it, use it as a starting point for  
a MCMC sampling method with fast mixing properties (such as HMC) 
to move to another point, record the sample, update $q$, and repeat the process. 
The samples from $q$ may suffer from having too large values of $\tilde{\mathcal{L}}_p$, 
so a Metropolis style acceptance rate can be added to prevent this. 
It is worth emphasizing that the philosophy 
of this paper is to find $q$ that fits the posterior everywhere: both 
samples from $q$ and samples from $p$ should lead to ${\rm EL_2O}$ close to zero. 
and if $q$ covers $p$ we should always find this solution in the limit of 
large sampling density. These issues will be explored further in 
next section, where we show for specific examples 
that more expressive $q$ reduce the 
differences between sampling from $q$ versus sampling from $p$ or $q+p$. 


\subsection{Hessian for nonlinear least squares and related loss functions}

One of the most common statistical analyses in science is a nonlinear 
least squares problem, which is a simple acyclic graphical 
model. One has some data vector $\bi{x}$ and some
model for the data $\bi{f}(\bi{z})$, which is nonlinear in terms of 
the parameters $\bi{z}$. We also assume a known data measurement noise covariance matrix $\bi{N}$, 
which can be dependent of the parameters $\bi{z}$. 
One may add a prior for the latent variables in the form, 
\begin{equation}
\tilde{\mathcal{L}}_p=\frac {1}{2} \left\{\bi{z}^T\bi{Z}^{-1}\bi{z}+[\bi{x}-\bi{f}(\bi{z})]^T\bi{N}^{-1}[\bi{x}-\bi{f}(\bi{z})]+\ln \det \bi{ZN}\right\},
\label{ls}
\end{equation}
where $\bi{Z}$ is the prior covariance matrix of $\bi{z}$ and we assumed the prior 
mean is zero (otherwise we also need to subtract out the prior mean from $\bi{z}$ 
in the first term). 

The Hessian in the Gauss-Newton approximation is
\begin{equation}
\E_{\tilde{p}}[\nabla_{\bi{z}} \nabla_{\bi{z}}\mathcal{L}_p] \approx 
 \bi{Z}^{-1}+ \E_{\tilde{p}}\left\{ 2{\rm tr} \left[\bi{N}^{-1} (\nabla_{\bi{z}}\bi{N}) 
\bi{N}^{-1}(\nabla_{\bi{z}}\bi{N})\right]+(\nabla_{\bi{z}}\bi{f})^T\bi{N}^{-1}(\nabla_{\bi{z}}\bi{f})
 \right\},
\end{equation}
where we dropped the second derivative terms of $\bi{f}$ and $\bi{N}$. The former 
term multiplies the residuals
$\bi{x}-\bi{f}(\bi{z})$, which close to the best fit (i.e. where the posterior mass 
is concentrated) are oscillating around zero 
if the model is a good fit to the data. This  
suppresses this term relative to the first derivative 
term, which is always positive: 
in the Gauss-Newton approximation the curvature matrix is explicitly positive definite, 
and so is its expectation value over the samples. Clearly this is no longer valid if we move 
away from the peak, or 
we have a multi-modal posterior, since we have extrema that are saddle points or local minima
and Gauss-Newton is a poor approximation there, 
so care must be exercised when using Gauss-Newton away 
from the global minimum. In our applications we use the Hessian to determine
$\bi{\Sigma}^{-1}$, but we ignore it when evaluating nonlinear parameters $\epsilon$ and $\eta$. 
The cost of evaluating the Hessian under the Gauss-Newton approximation 
equals the cost of evaluating the gradient $\nabla_{\bi{z}}\mathcal{L}_p$, since it simply involves
first derivatives of $\bi{f}$ or $\bi{N}$. 

\subsection{Range constraints}

If a variable has a boundary then finding a function extremum 
may not be obtained by finding where its gradient is zero, but may instead be
found at the boundary. In this case the posterior distribution is abruptly 
changed at the boundary, which is difficult to handle with 
Gaussians. 
The most common case is that a given variable is bounded to a one sided 
interval, or sometimes to a two-sided interval.
There are two methods one can adopt, first one is a transformation to 
an unconstrained variable and second one is a reflective boundary condition. 

Suppose for example that we have a constraint 
$z_i'>a_i$, and we would like to have an 
unconstrained optimization that also transitions to the $z'$ prior on a scale  $\xi_i$ away from $a_i$.
We can use
\begin{equation}
z_i=\xi_i\ln\left(e^{\frac{z_i'-a_i}{\xi_i}}-1\right)
\label{uncon}
\end{equation}
as our new variable \citep{KucukelbirTRGB17}. 
This variable becomes $z_i'$ for $z_i'\gg a_i+\xi_i$ and $\xi_i\ln (z_i'-a_i)/\xi_i$ for
$a_i<z_i'\ll a_i+\xi_i$, so $z_i$ is now defined on the 
entire real interval with no constraint. If we want to preserve the probability and use 
the prior on original $z'$ we must also 
include the Jacobian $J_i=|dz_i/dz_i'|$, $p(z_i)=p(z_i')J_i^{-1}$.  
The presence of the Jacobian modifies the loss 
function.
In our examples below we are including the Jacobian and we treat $\xi_i$
as another nonlinear parameter attached to parameter $z_i$ with a 
constraint, so that we optimize EL$_2$O with respect to it. 
A problem with this method is that posterior will always go to zero 
at $z_i' \rightarrow a_i$, since the Gaussian goes to zero
at large values. So even though this method can be quite 
successful in getting most of the posterior correct, it 
will artificially turn down to zero 
at the boundary. This is problematic, 
since it suggests the data exclude the boundary even if they 
do not.

Second approach to a boundary $z_i'>a_i$ is to extend the 
range to $z_i'<a_i$ using a reflective (or mirror) boundary condition across 
$z_i'=a_i$, such 
that if $z_i'<a_i$ then $\tilde{\mathcal{L}}_p(z_i'-a_i)=\tilde{\mathcal{L}}_p(a_i-z_i')$.
This leads to the non-bijective transformation of 
appendix A with $b_i=0$: we have $z_i'$ defined on 
entire range and we model it with a sum of two mirrored Gaussians. 
Effectively this is 
equivalent to an unconstrained posterior analysis, where 
we take the posterior 
at $z_i'<a_i$ and add it to $z_i'>a_i$.
It solves the problem of the unconstrained
transformation 
method above, as the posterior at the boundary is not forced to zero, 
since it can be continuous and non-zero across the boundary $a_i$. The marginalization 
over this parameter remains trivial, since it is as if the parameter is not constrained at all. 
For the purpose of the marginalized posterior for the 
parameter itself, we must add the $z_i'<a_i$ 
posterior to $z_i'>a_i$ posterior. If the posterior mass is non-zero at 
$z_i'=a_i$ then this will result in the posterior abruptly transitioning
from a finite value to 0 at the boundary. 
This method can be generalized to a two sided boundary. 
In section \ref{sec4} we will show an example of both 
methods.

\subsection{Related Work}
Our proposed divergence is in the family of f-divergences. Recently, several divergences have been introduced 
(e.g. \citet{RanganathTAB16,DiengTRPB17})
to counter the claimed problems of KL divergence such as its asymmetry and exclusivity of $q$, but here
we argue that with 
sufficiently expressive $q$ these problems may not be fundamental 
for EL$_2$O method. Expectations of EL$_2$O equations agree with VI 
expressions of 
\citep{OpperArchambeau09}. 

Stochastic VI has been explored for posteriors in several papers, 
including ADVI \cite{KucukelbirTRGB17}. In direct comparison test we 
find it has a slower convergence than EL$_2$O.
For $n=1$ the Fisher divergence minimization has been proposed by \citet{Hyvarinen05} as a 
score matching statistic, but was rewritten 
through integration by parts into a form that does not cancel sampling variance and has similar convergence properties as stochastic VI.
 Reducing sampling noise has also been explored more recently in 
\citet{RoederWD17} in a different context and with a different approach. 
Quantifying the error of the VI approximation has been explored in \citet{YaoVSG18}, but  
using EL$_2$O value
is simpler to evaluate. 

NL transformations have been explored in terms of boundary effects in \citet{KucukelbirTRGB17}. Our NL transformations
correspond more explicitly to generalized skewness and curtosis parameters, 
and as such are useful for general description of probability distributions. We employ 
analytic marginals to obtain posteriors and for this reason we only employ a single layer point-wise NL transformations, instead of the more powerful 
normalizing flows \cite{RezendeMohamed15}. More recently, \cite{LinKS19}
also adopt GM and NL for similar purposes, also using Hessian based 
second order optimization (called natural gradient in recent ML literature).

\section{Numerical experiments}
\label{sec4}

In this section we look at several examples in increasing order of complexity. 

\subsection{Non-Gaussian correlated 2D posterior}

\begin{figure}[t!]
\centering \includegraphics[height=0.34\textwidth]{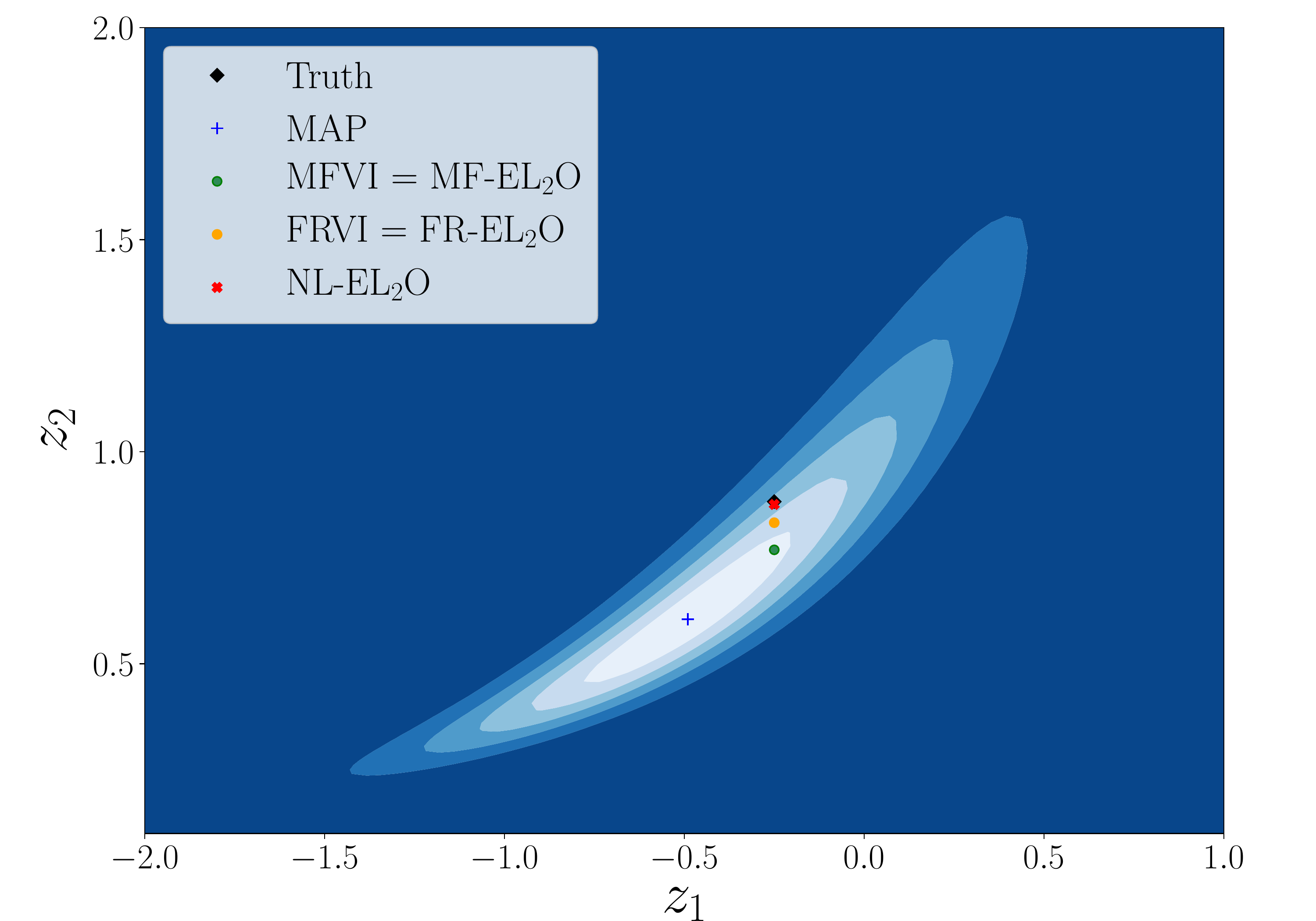} \hspace{0cm} \includegraphics[height=0.34\textwidth]{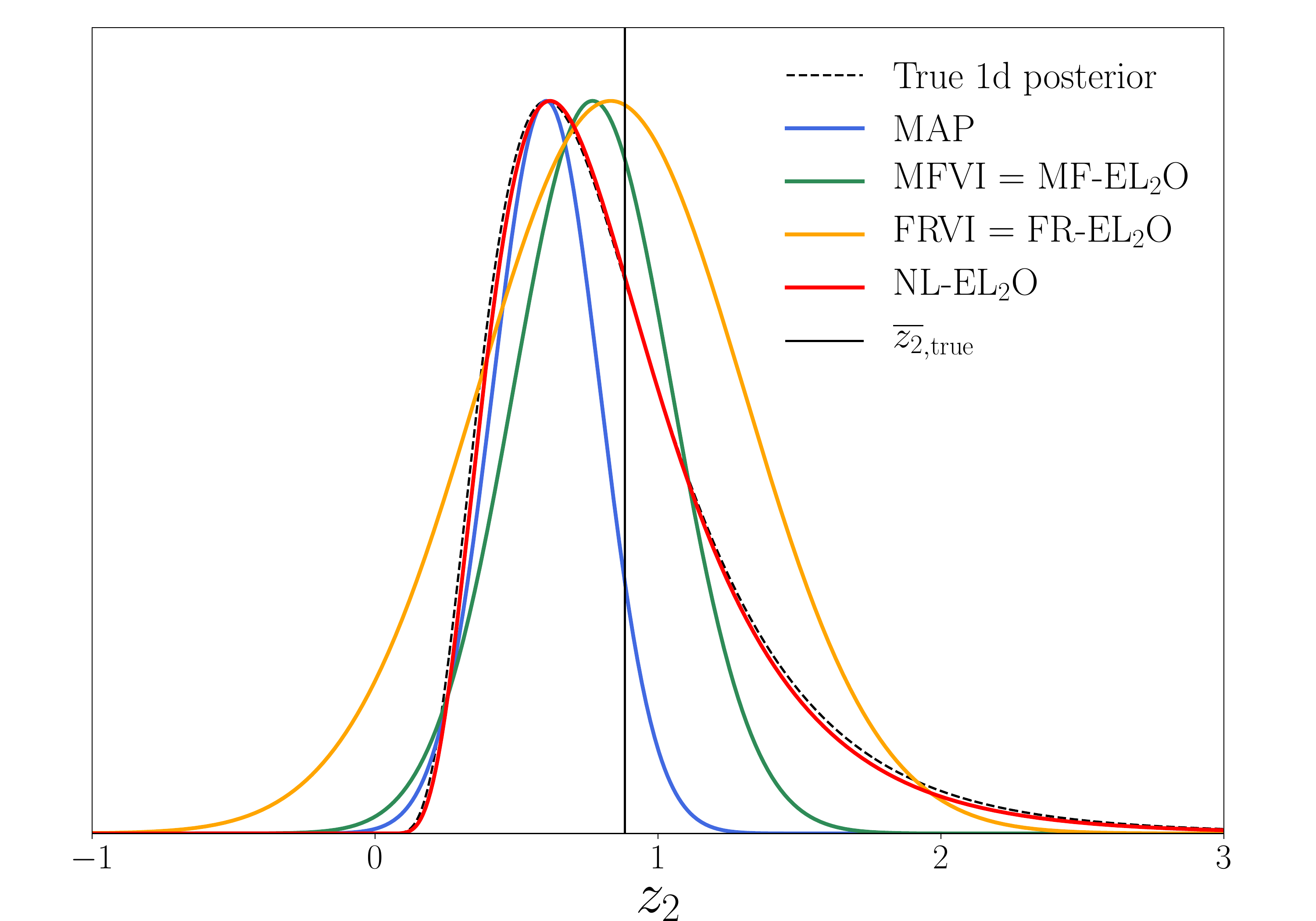}
\caption{Example of a correlated non-Gaussian posterior problem, where one of the two Gaussian correlated variables $y_2$ is preceded by a nonlinear transformation (mapped by the exponential function, $z_2 = \mathrm{exp}(y_2)$). \textit{Left}: The 2D posterior and the means estimated by various methods. \textit{Right}: 1D marginalized posterior of $z_2$, with the black vertical line marking its true mean. MAP (blue) finds the mode and MFVI (green)
FRVI (yellow) estimate the mean relatively well, but all of them fail to capture the correct shape of the posterior and its variance. Fitting for the skewness and curtosis parameters, EL$_2$O with the NL transform (NL-EL$_2$O, red) accurately models the posterior. All curves have been normalized to the same value at the peak to reduce their 
dynamical range.}
\label{banana}
\end{figure}

\begin{figure}[b!]
\centering \includegraphics[height=0.34\textwidth]{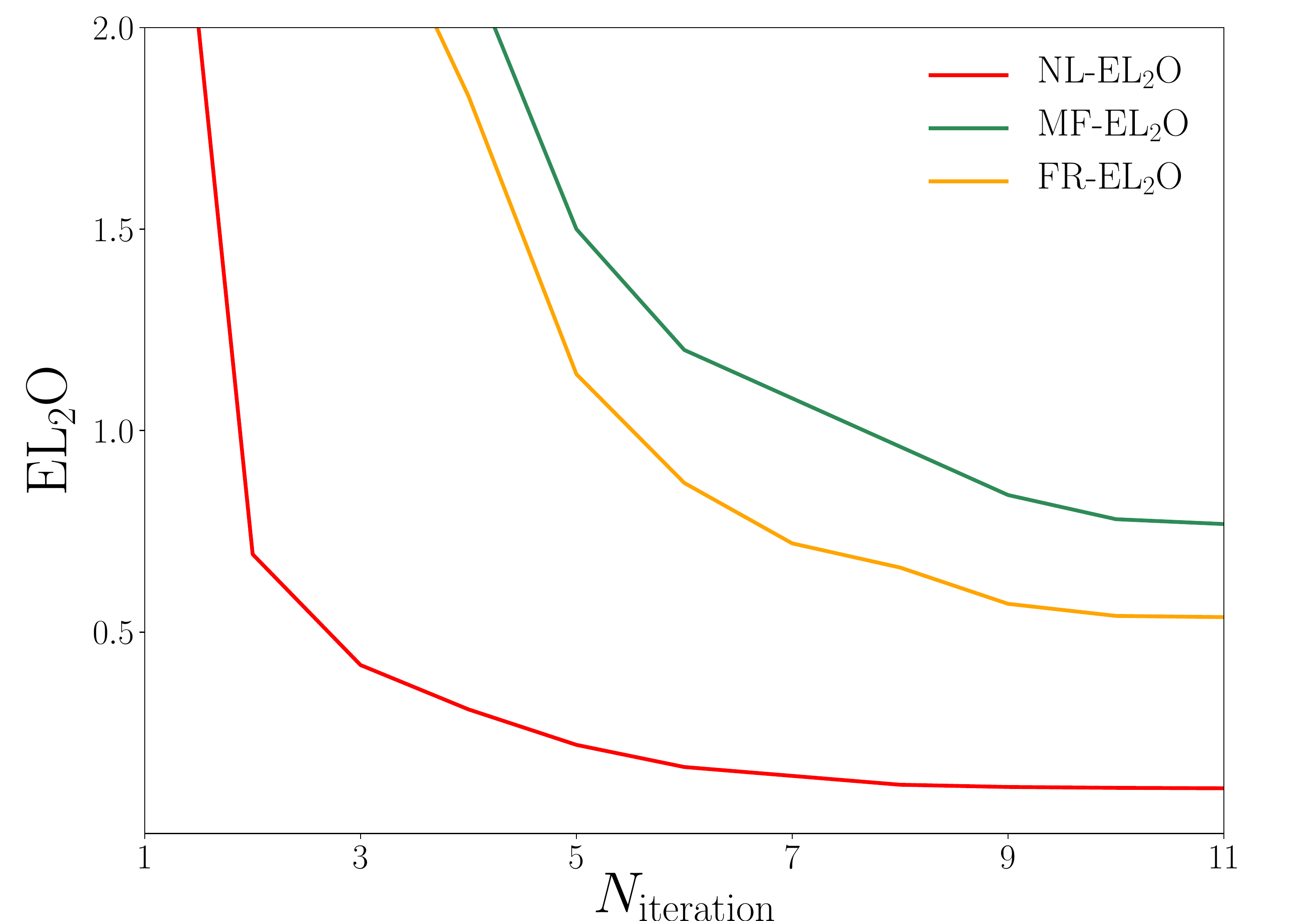} \includegraphics[height=0.34\textwidth]{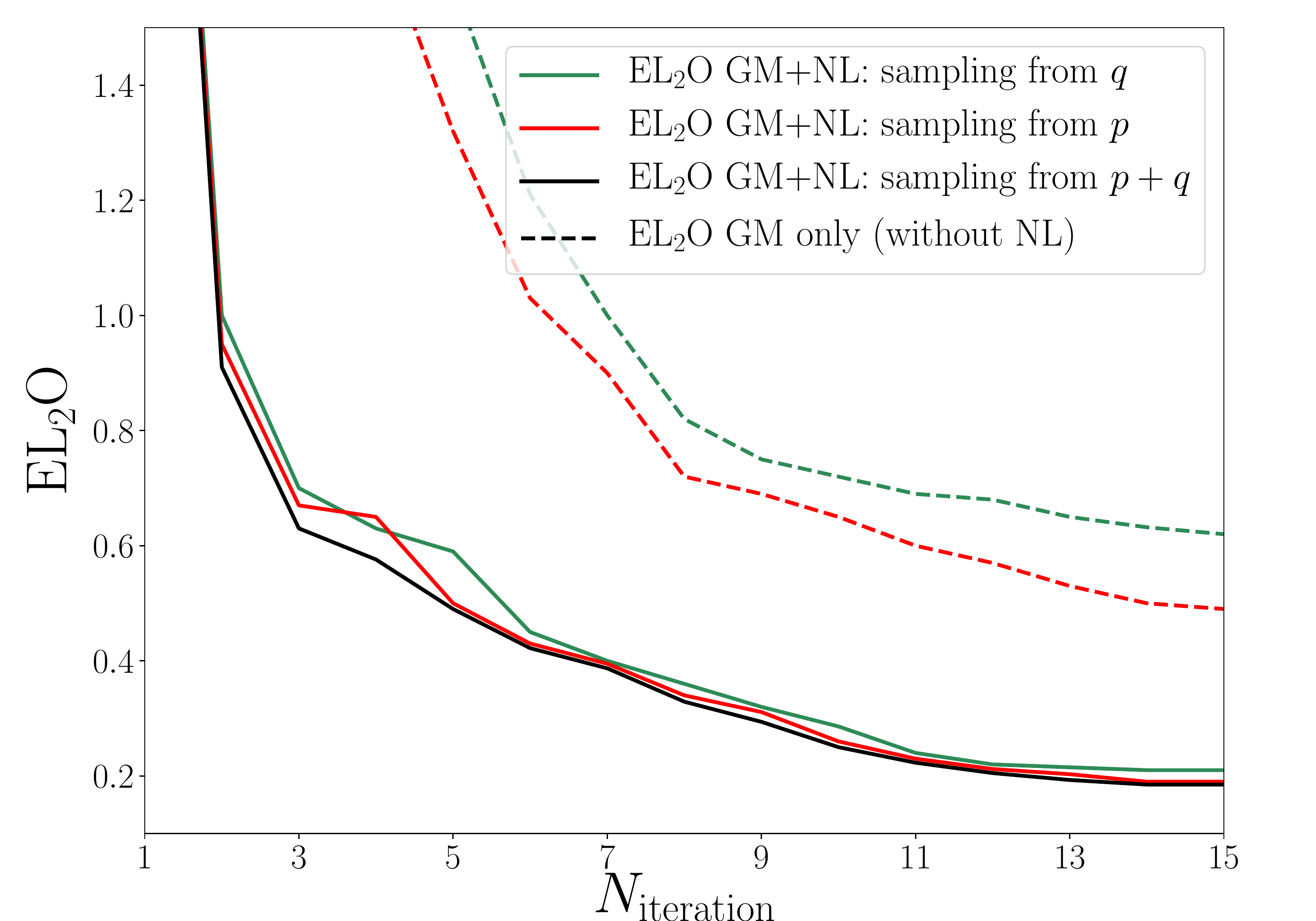}
\caption{EL$_2$O values provide an estimate of the quality of the fit. Here we show
them as a function of the number of iterations for the correlated non-Gaussian posterior example (\textit{Left panel}) and for the forward model posterior example (\textit{Right panel}). Typically values of EL$_2$O $\lesssim 0.2$ indicate that we have obtained a 
satisfactory posterior. The convergence is faster for NL-EL$_2$O despite 
having more parameters, a consequence of sampling noise free 
nature of EL$_2$O. In these examples each iteration draws 5 samples and we average 
over the past samples after the burn in. \textit{Right}: Sampling from $p$ (red dashed) gives better EL$_2$O that sampling from $q$ (green dashed) for full rank GM. As we increase the expressivity of $q$, by applying the NL transform to better match the posterior, this makes EL$_2$O values for sampling from $p$ and $q$ more similar, and lower, improving the overall quality of the posterior.}
\label{ELO}
\end{figure}

In the first example we have a 2-dimensional problem modeled as two Gaussian distributed 
and correlated variables $z_1$ and $z_2$, but second one is 
nonlinear transformed using $\exp(z_2)$ mapping. 
This transformation is not in the family of 
skewness and curtosis transformations proposed in section \ref{sec3}. Here we will try to model the 
posterior using $\epsilon$ and $\eta$ in addition to $\bi{\mu}$ and $\bi{\Sigma}$. The question 
is how well can our method handle the posterior of $z_2$, as well as the joint posterior 
of $z_1$ and $z_2$, and how does it compare to MAP, MF and FR VI or EL$_2$O. 

The results are shown in figure \ref{banana}. Left panel shows the 2D contours, 
which open up towards larger values of $z_2$ and as a result the MAP is away from the 
mean. Right hand panel shows the resulting posterior of MAP+Laplace, MF and FR EL$_2$O
(which equals MFVI and FRVI in large sampling limit), and 
NL EL$_2$O. MAP gets the peak posterior correct but not the mean, MF improves on the 
mean and FR improves it further, but none of these get the full posterior. NL EL$_2$O
gets the full posterior in nearly perfect agreement with the correct distribution, which is , 0.13 versus 0.5 or 0.7, 
respectively. What is interesting that the convergence of NL EL$_2$O is faster, 
despite having more parameters: the convergence has been reached 
after 8 iterations. We started with $N_k=1$ and ended with $N_k=5$ for this example. Note that we can reuse 
samples from previous iterations. 

\subsection{Forward model posterior}

\begin{figure}[t!]
\centering \includegraphics[height=0.34\textwidth]{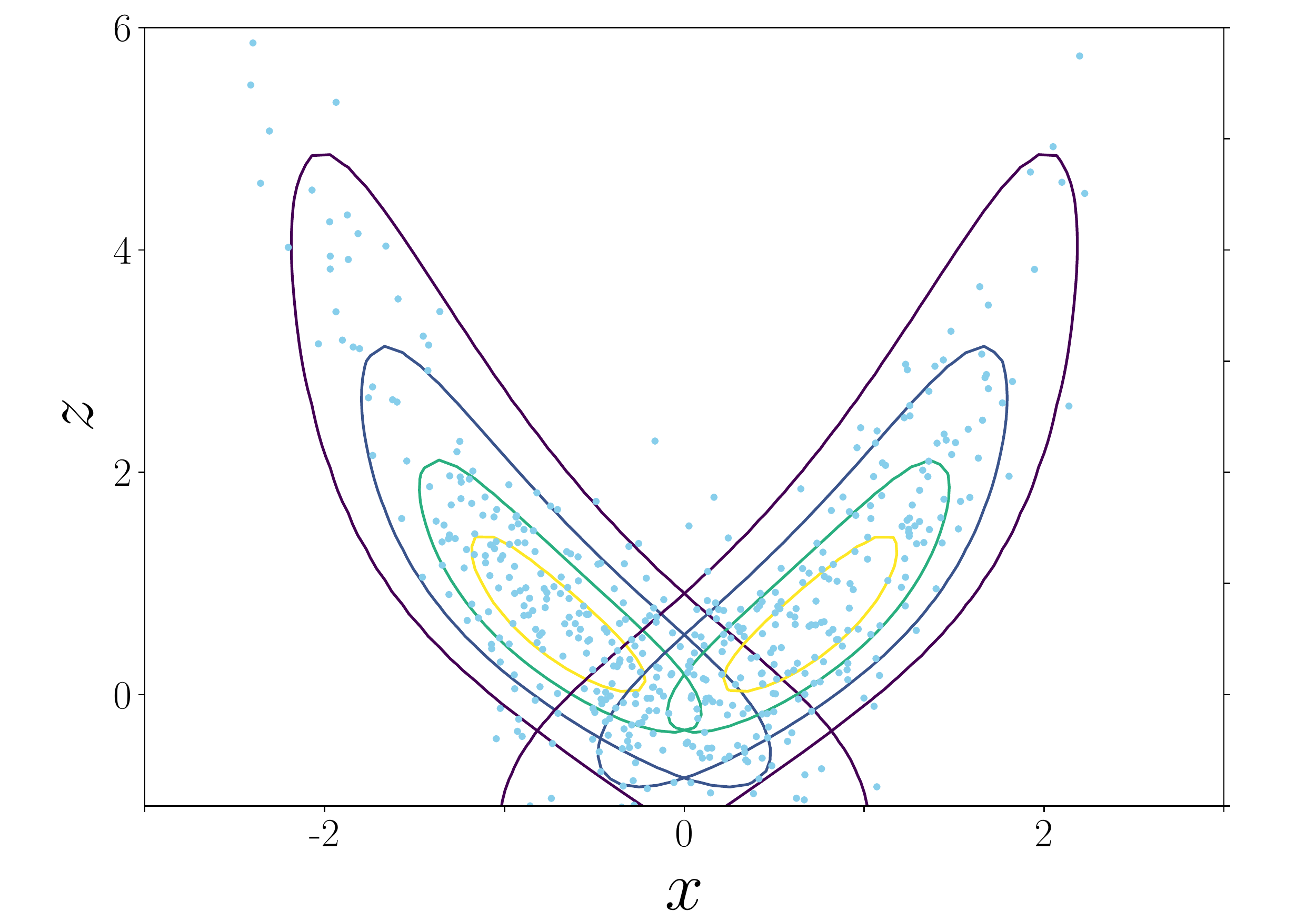} \hspace{0cm} \includegraphics[height=0.34\textwidth]{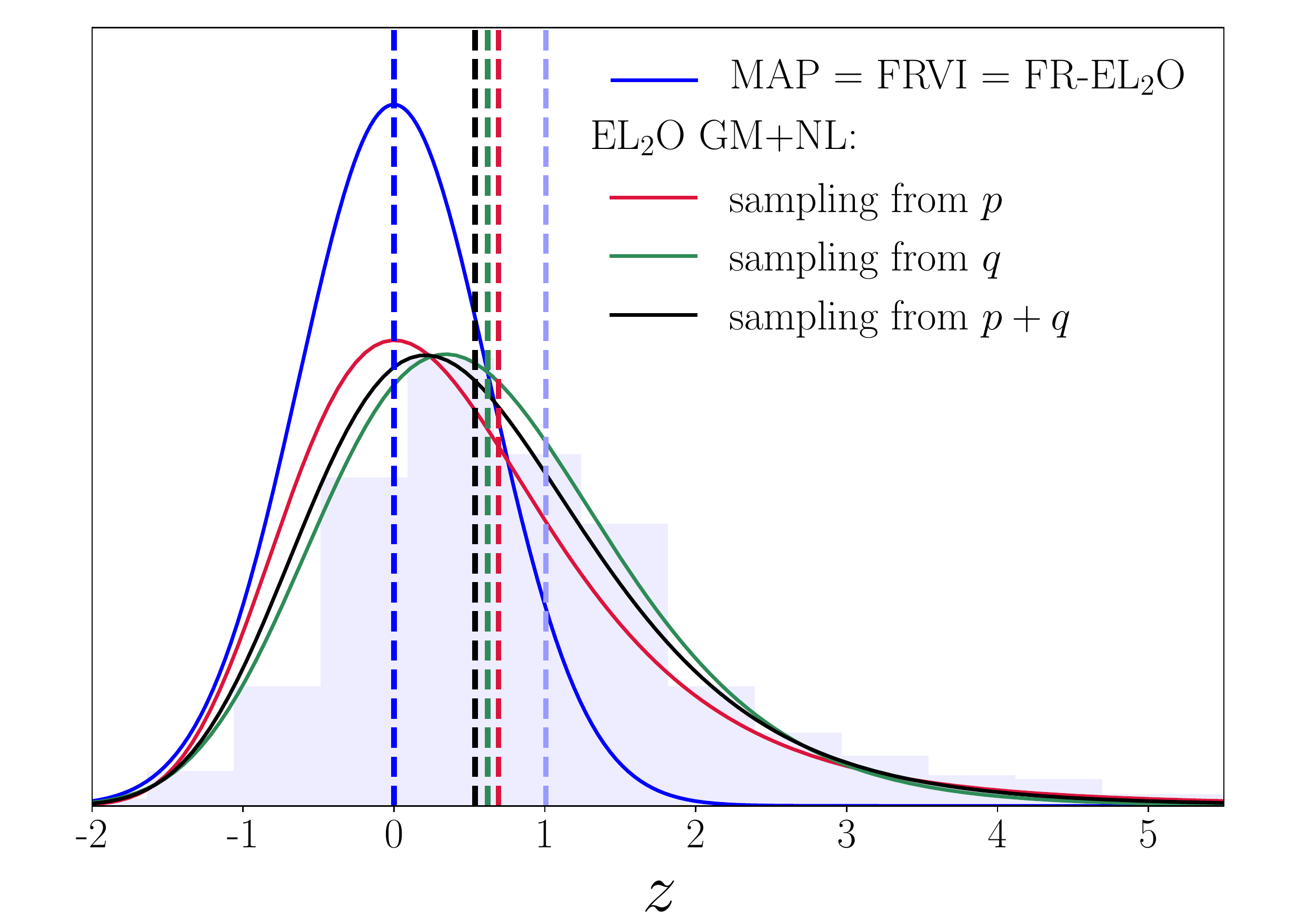}
\caption{Example of a forward inference problem. 
\textit{Left}: Contours of two symmetric Gaussian components (GM), with NL transform applied, together with samples from the posterior. The elliptical contours are warped by NL transform to better match the posterior. The total 
posterior is the sum of the two, which enhances the posterior density at $x\sim 0$.
\textit{Right}: 1D marginalized posterior of $z$ as approximated by different methods. MAP and FRVI (blue, normalized to the same peak value to reduce the dynamic range of the plot) give a poor estimate of posterior compared to MCMC (histogram). For EL$_2$O, we  evaluated GM+NL with the following sampling proposals $\tilde{p}$: sampling from $p$ (red), $q$ (green), and $p+q$ (black). Note that sampling 
from $p$ is narrower than sampling from $q$. 
Vertical bars indicate the means, including MCMC (light blue dashed). 
}
\label{pq}
\end{figure}

A very simple state evolution model
is where we know 
the prior distribution of $x$, assumed to be a Gaussian with zero mean and 
variance $\Sigma$, and we would like to 
know the posterior of $z=x^2+n$, where 
$n$ is Gaussian noise with zero mean and variance $Q$. 
The loss function $\tilde{\mathcal{L}}_p$ is
\begin{equation}
\tilde{\mathcal{L}}_p=\frac{1}{2}\left[x\Sigma^{-1}x
+(z-x^2)Q^{-1}(z-x^2)\right],
\label{likelihood}
\end{equation}
where we dropped all irrelevant constants. 
We would like to find the posterior of $z$ marginalized over $x$. 
In the absence of noise the problem can be solved using the Jacobian between $x$ and $z$, 
but addition of noise requires an additional convolution. In higher dimensions evaluating 
the Jacobian quickly becomes very expensive, so we will instead solve the problem by 
approximating the joint probability distribution of $x$ and $z$, and then 
marginalizing over $x$. This is a hard problem because the joint 
distribution is very non-Gaussian, as seen in figure \ref{pq}. 

We can first attempt to solve with MAP. The MAP solution is at $\hat{x}=\hat{z}=0$, 
and at the MAP Laplace approximation gives a diagonal Hessian between $x$ and 
$z$, so the two are uncorrelated. The variance on $z$ is $Q$, which 
vanishes in no noise $Q \rightarrow 0$ limit. MAP+Laplace for $z$ is thus 
a narrow distribution at zero, which is clearly a very 
poor approximation to the correct posterior. 

The full rank VI or ${\rm EL_2O}$ approach is to sample from full rank Gaussian 
$q$ and iterate 
until convergence. The off-diagonal elements of the Hessian are given by 
$\nabla_z\nabla_x\mathcal{L}_p=-2xQ^{-1}$, which vanishes upon 
averaging over $x$, so full rank and mean field solutions are equal
and FRVI assumes the two variables are uncorrelated. The variance on 
$z$ is again given by $Q$.
The inverse variance of $x$ is $\nabla_x\nabla_x\mathcal{L}_p=\Sigma^{-1}+4x^2Q^{-1}$, 
which in ${\rm EL_2O}$ or VI we need to average over $q$. 
The stationary point is reached when $\E[ \nabla_x\nabla_x\mathcal{L}_p]^{-1}=\E[ x^2 ]=\sigma_x^2$, which in the low $Q$ limit 
gives $\sigma_x^2=Q^{1/2}/2$. Once we have the full rank Hessian we can also determine 
the means from the gradient of equation \ref{likelihood}, finding a solution $\mu_x=\mu_z=0$. 
So we find a somewhat absurd result that even though 
$x$ is not affected by $z$ and its posterior should equal its prior, 
FRVI gives a different solution, one of a delta function at zero in the $Q \rightarrow 0$ limit, 
which is identical to the MAP solution. 
However, with this posterior the value of ${\rm EL_2O}$ is large, because the Hessian 
is fluctuating across the posterior and is not well represented with a single average. 
This is specially clear for the off-diagonal elements, whose average is zero, but the actual 
values fluctuate with rms of order $Q^{-3/4}$, very large 
fluctuations if $Q \rightarrow 0$. 

One must improve the model by going beyond a single full rank Gaussian.
 Here we will do so with a non-bijective transformation of 
appendix A, using $b_x=0$ and $a_x=0$, i.e. we model it as 
two Gaussian components
mirror symmetric across $x=0$ axis. We use equation \ref{hessgm}, which says that if the two Gaussian 
components are well separated
the local Hessian can be used to determine $\bi{\Sigma}^{-1}$ of the local 
Gaussian component (which then also determines the covariance of the other component 
due to the symmetry). 
The local Hessian 
is given by $\nabla_x\nabla_x\mathcal{L}_p=\Sigma^{-1}+4\mu_x^2Q^{-1}$, 
$\nabla_x\nabla_z\mathcal{L}_p=-2\mu_xQ^{-1}$ and   $\nabla_z\nabla_z\mathcal{L}_p=Q^{-1}$.

To further improve the model we consider bijective 
nonlinear transforms (NL). 
These are useful as they warp the 
ellipses, which allows to match $q$ closer to the true 
posterior $p$. 
The results are shown in figure \ref{pq}. We see from the figure that MAP or FRVI=FR-EL$_2$O 
fail to give the correct posterior, while the Gaussian mixture with NL gives 
a very good posterior of $z$, in good agreement with MCMC. 

In this example we can sample from $p$ directly, so we do not need to iterate 
on samples from $q$. 
${\rm EL_2O}$ has flexibility to use samples from either of the two, which is distinctly 
different from KL based VI. This forward model problem 
gives us the opportunity to compare
the results between the two. 
We would like to know if 
sampling from $q$ versus $p$ gives different 
answers, and if sampling from both further improves 
the results. This is also shown in figure \ref{pq}. We see that there are some 
small differences in the posteriors, and that sampling from $q$ is slightly worse: 
in terms of ${\rm EL_2O}$ value, we get 0.20 for sampling from $p$ and 
$p+q$ and 0.23 for sampling from $q$. Somewhat surprising, we find that 
sampling from $q$ gives a broader approximation 
that sampling from $p$, contrary to KL divergence based FRVI
\citep{Bishop07}.
Sampling from $p+q$ does 
not further improve the results over sampling from $p$. 
The difference between 
$p$ and $q$ sampling is
larger if we restrict to the full rank Gaussian without 
NL, and the ${\rm EL_2O}$ values are also larger: 0.4 for $p$ versus 0.5 for $q$. 
This suggests that while for simple $q$ the results may 
be biased and sampling from $p$ is preferred, more expressive
$q$ reduces the difference between the two. This is not 
surprising: if $p$ is in the family of $q$ then we should be 
able to recover the exact solution with 
optimization, finding ${\rm EL_2O}=0$ upon convergence. 
If we want to improve the $q$ sampling 
results of figure \ref{pq} we can do so 
by adding additional Gaussian mixture components, or 
additional NL transforms, but we have not attempted to do so here.

A potential concern is that the exclusive nature of $q$ may lead to a situation 
where EL$_2$O sampled from $q$ is low, but the quality is poor, because $p>0$
where $q=0$. If this happens because there is another posterior maximum elsewhere
far away then the only way to address it is using global optimization techniques
like multiple starting points. All methods, including MCMC, have difficulties in these
situations and require specialized methods (see below). If, on the other hand, there is excess
posterior mass that is smoothly attached to the bulk of $q$ then EL$_2$O method
should be able to detect it, specially with gradient and Hessian information. 
We can test it on this example 
by comparing EL$_2$O values on samples evaluated from $p$ versus samples evaluated
from $q$, while using the same $\mathcal{L}_q$ to evaluate EL$_2$O. 
We find very little difference between the two, 0.24 versus 0.23, and so 
${\rm EL_2O}$ evaluated on samples from $q$ gives a 
reliable estimate of the quality of solution. 
In these examples we used up to
15 iterations with $N_k=5$, and 
averaging over all past iterations after the burn-in. 

\subsection{Multi-modal posterior}

Multi-modal posteriors are very challenging for any method. If the 
modes are widely separated then standard MCMC methods will fail, and specialized 
techniques, such as annealing or nested sampling \citep{HandleyHobsonEtAl15} are required.
If the modes are closer to each other so that their posteriors overlap 
then MCMC will be able to find them, and 
this is considered to be a strength of MCMC as compared to MAP or VI, which 
in the simplest implementations find only one of the modes. Here we 
use EL$_2$O with a Gaussian mixture (GM) on a simple bimodal posterior, which is a sum of two 
full rank Gaussians in 2 dimensions. 

\begin{figure}[t!]
\centering \includegraphics[height=0.32\textwidth]{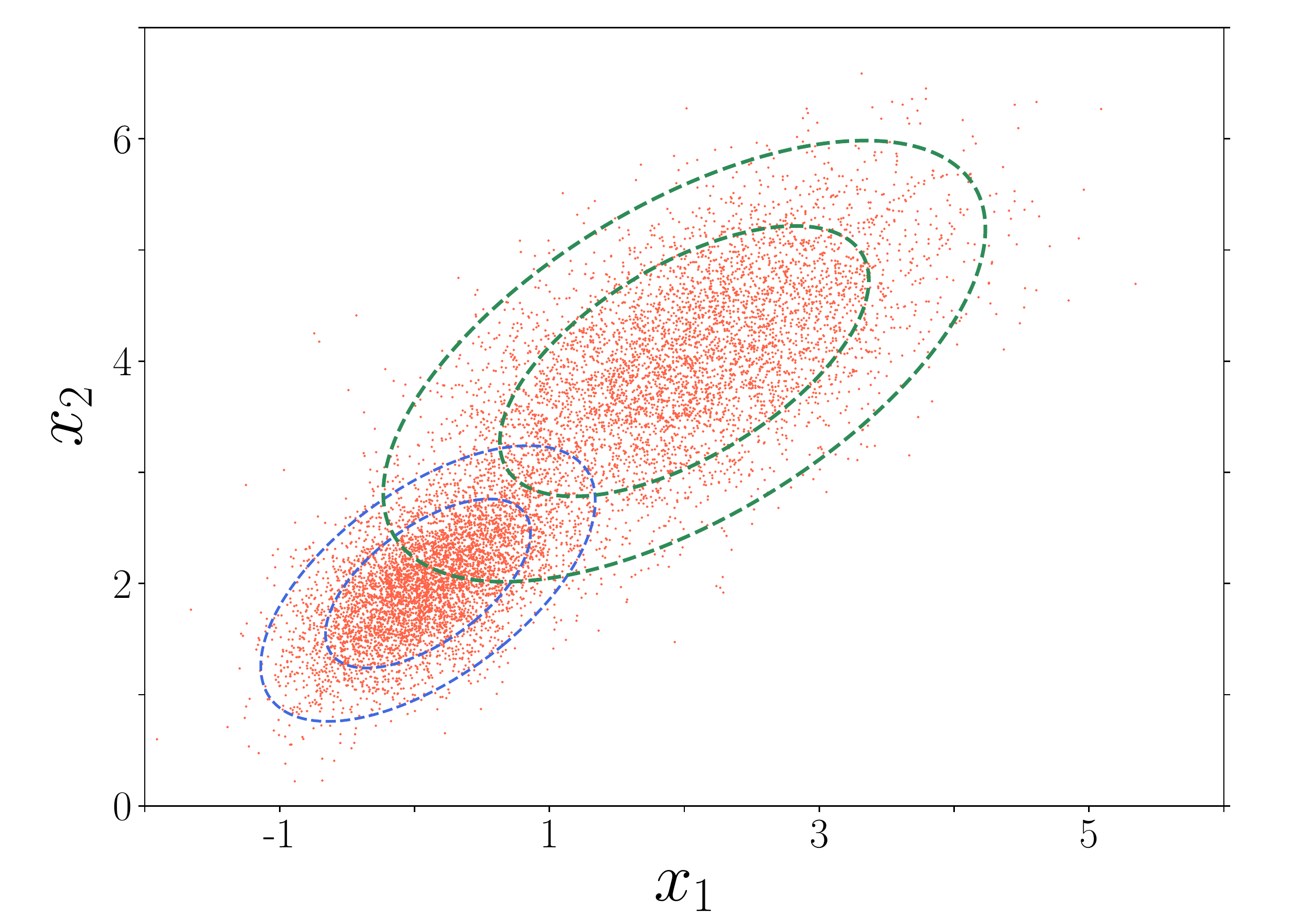} \includegraphics[height=0.32\textwidth]{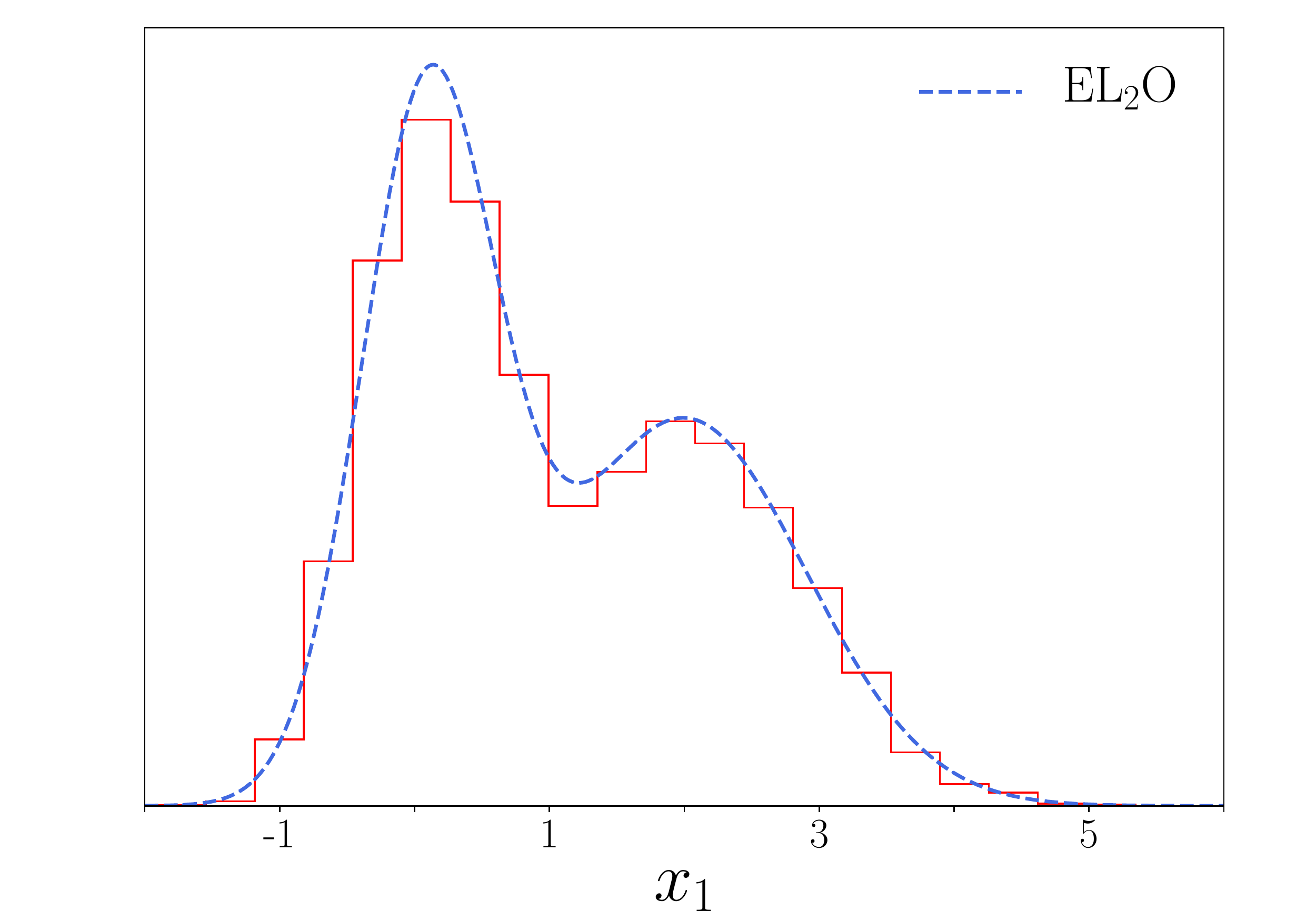}
\caption{Application of a Gaussian mixture (GM) model to the multi-modal posterior problem. 
\textit{Left}: Modeling the posterior as a weighted sum of two bivariate Gaussians, we demonstrate that the EL$_2$O method identifies both peaks, with means and covariances accurately estimated. A single starting point with 2 GM components converges to this solution after 15 iterations. For multiple starting points, each one converges within a few iterations to one of the two local minima, and EL$_2$O properly normalizes the two GM components. The two final solutions are identical (we show the multiple starting point method). 
\textit{Right}: 1D marginalized posterior predicted by the EL$_2$O (blue dotted line), which closely matches the posterior from samples (red).}
\label{fig:multimodal}
\end{figure}

We show the results in figure \ref{fig:multimodal}. 
For this example we consider two optimization strategies. The 
first one is to first iterate on a full rank Gaussian, 
and since the residuals are large, we add a nonlinear transform. 
Since even after this residuals remain large we add a second 
Gaussian component. After a total of 13 iterations we converge to the correct 
posterior. This can be compared to Stein discrepancy 
method of 
\cite{LiuWang18}, where 500 iterations with 100 particles
 were used to converge. 
The convergence to the correct 
result is possible because the two 
modes overlap in their posterior density. 

The second strategy for these problems is to have multiple starting points. We will not discuss strategies how to choose 
the starting points, and we will adopt a simple random starting 
point method. 
In figure \ref{fig:multimodal} we show results with 
several different starting points, each converging within a few iterations to one of two two modes (about half of the time onto each). How many starting points we need to choose depends 
on how many modes we discover: if after a few starting points we 
do not discover new modes we may stop the procedure. 
We construct the initial solution as the sum of the two 
Gaussians as found at each mode, using the gradient and Hessian 
to determine the full rank Gaussian, with the relative normalization determined by equation \ref{gradfree}.  
If the two modes are widely separated this is already 
the correct solution, but in this case they are not and
we 
use optimization to further improve 
on the initial parameters. The end result
was identical to the above strategy, but the multiple starting points strategy is more robust, as it will
find a solution even for the case of widely separated modes. In general, 
if multimodal posteriors are suspected (and even if not), multiple starting points 
are always recommended as a way to verify that optimization found all the relevant
posterior peaks. 

\subsection{A science graphical model example: galaxy clustering analysis}

\begin{figure}[t!]
\centering \includegraphics[height=0.47\textwidth]{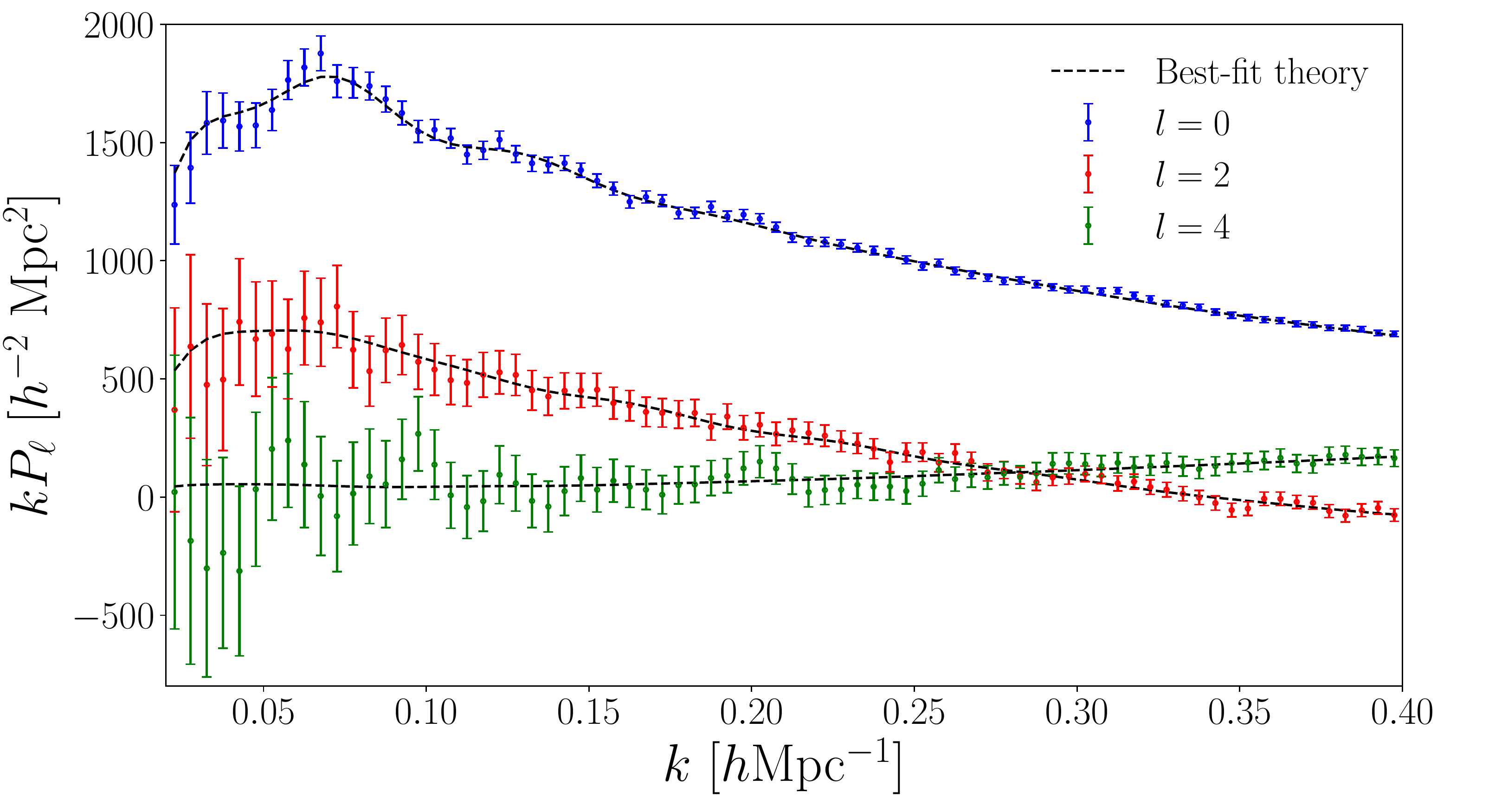}
\label{fig:Pk1}
\caption{The power spectrum multipoles ($l = 0, 2, 4$) from the best-fit theory model and measurements from the BOSS DR12 LOWZ+CMASS NGC data, with $0.4 < z < 0.6$.  Fitting the model to data over the wavenumber range $k = 0.02 - 0.4 h$Mpc$^{-1}$, we find a good agreement between the model and the measurements.}
\end{figure}

\begin{figure}[t!]
\centering
 \includegraphics[height=0.45\textwidth]{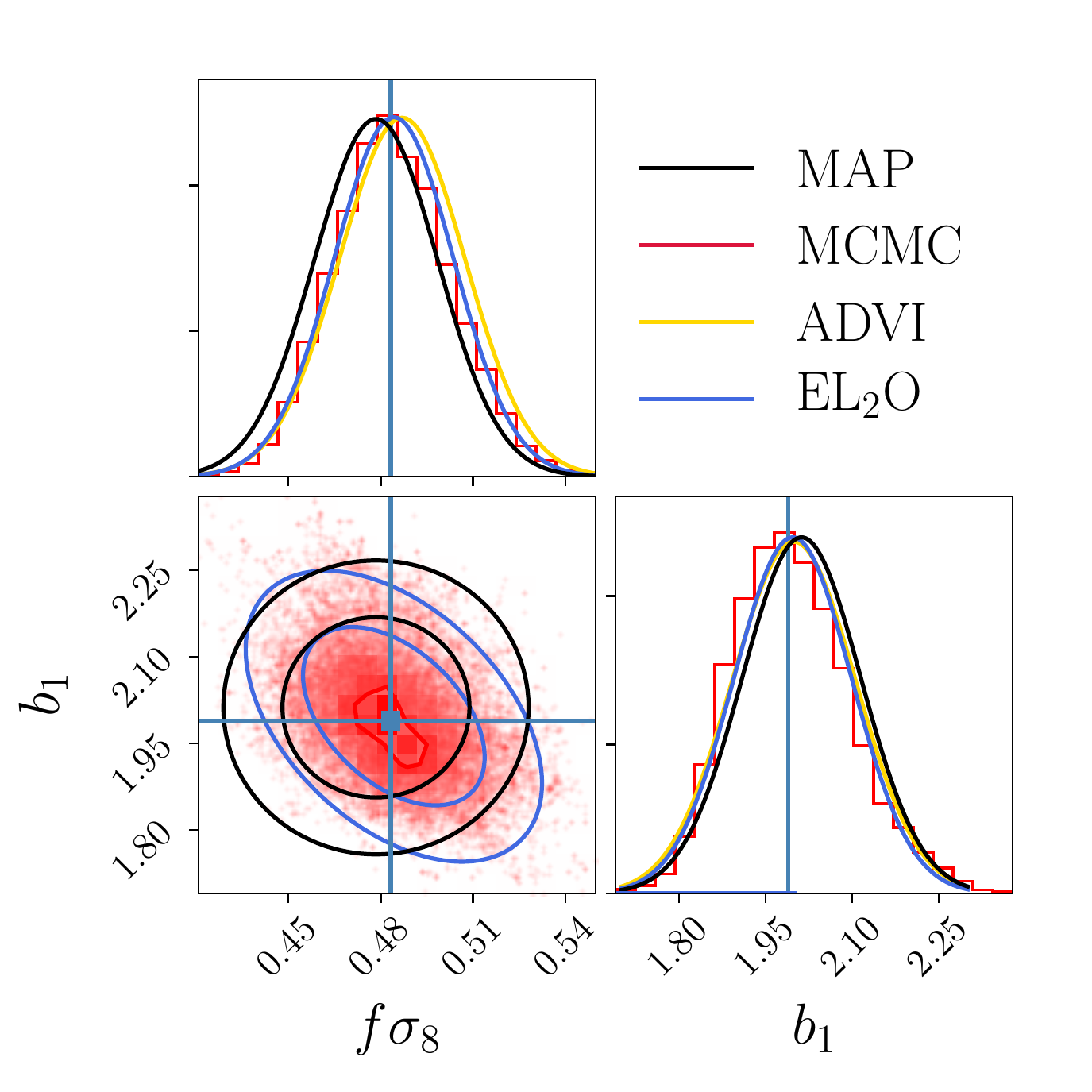} \includegraphics[height=0.45\textwidth]{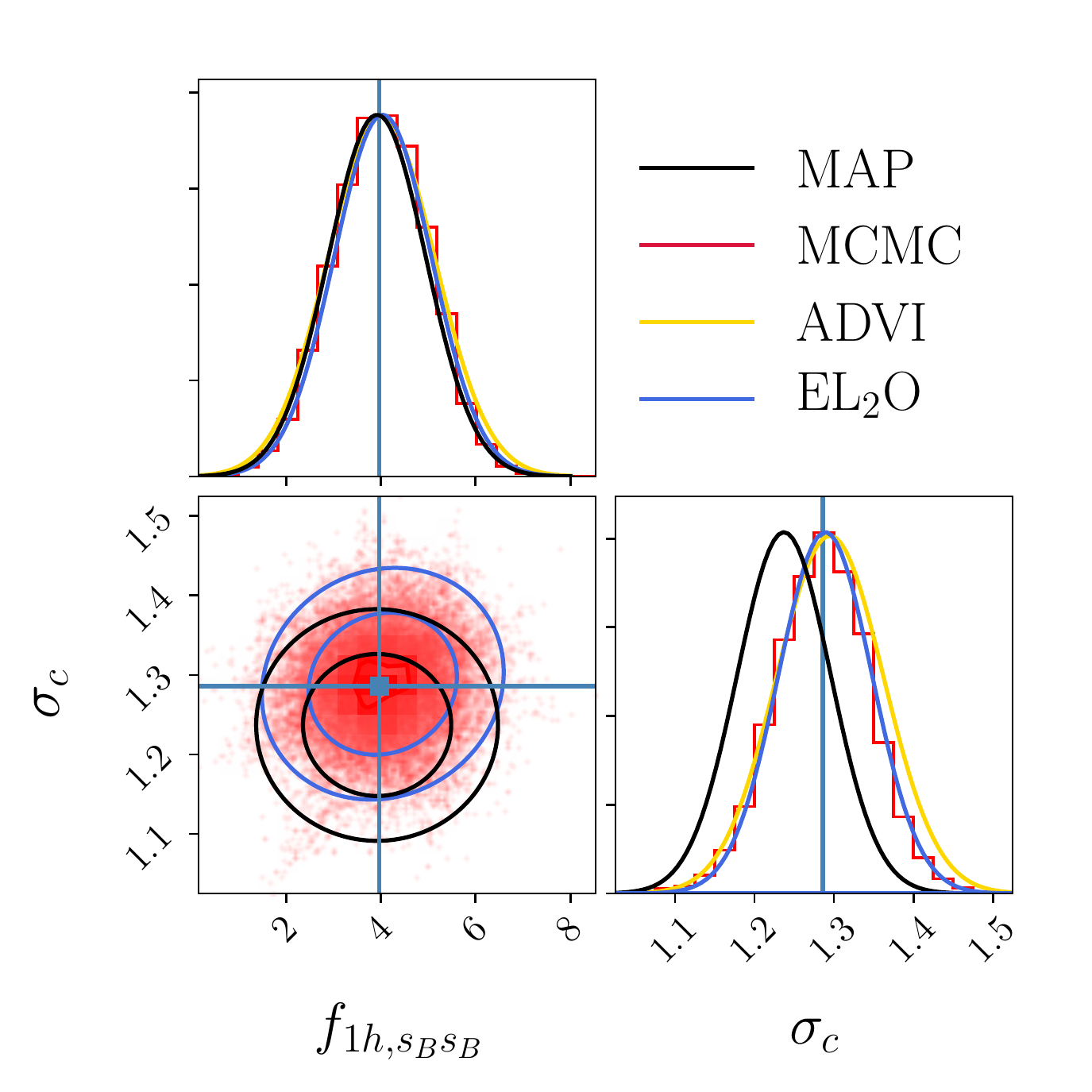} \includegraphics[height=0.25\textwidth]{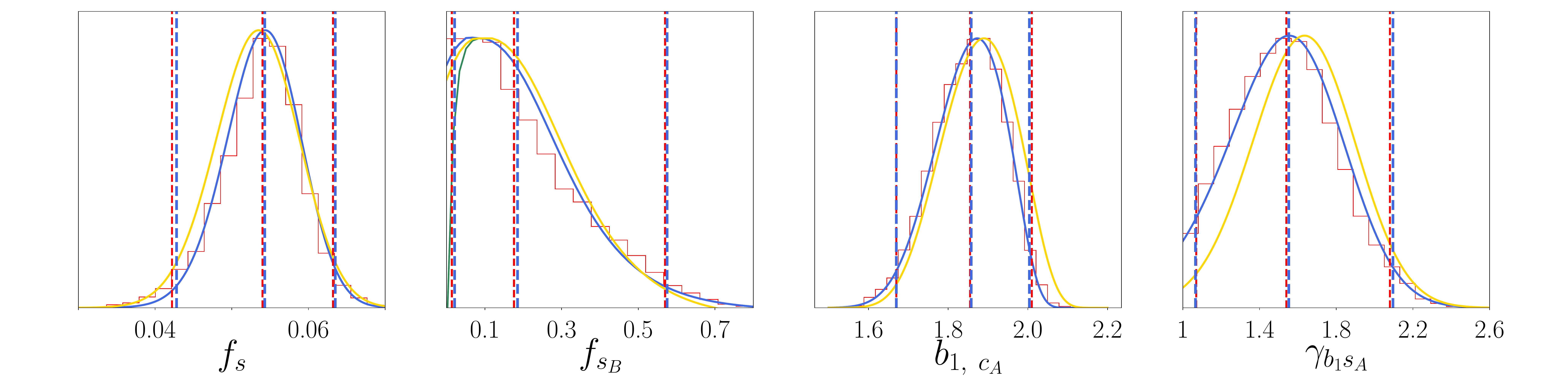}
\caption{\textit{Top}: 1D and 2D posterior distributions of four selected RSD model parameters whose posteriors are close to Gaussian. 
\textit{Top left panel}: MAP+Laplace gives inaccurate 2D posterior relative to EL$_2$O, even if 1D projections 
are accurate. \textit{Top right panel}: MAP can be displaced in the mean, while EL$_2$O and ADVI 
results agree very well with MCMC samples. \textit{Bottom}: 1D posteriors for parameters which are most non-Gaussian.
Together with the NL transform (blue solid curves), EL$_2$O results closely match the MCMC posterior (red solid). Also shown are 2.5\%, 50\%, and 97.5\% intervals (dotted lines), for MCMC and EL$_2$O. 125 likelihood evaluations  
were used for EL$_2$O, compared to $10^5$ for MCMC, and $2.3\times 10^4$ for ADVI. Despite taking 200 times more steps than ${\rm EL_2O}$, ADVI posteriors are 
considerably worse. 
For 
$f_{sB}$ parameter we have a boundary $f_{sB}>0$, and we 
model it with the unconstrained transformation method (green solid) and 
adding the
reflective boundary method to it, the latter allowing
the posterior density at the boundary to be non-zero (blue solid).}
\label{fig:Pk2}
\end{figure}

Our main goal is a fast determination of posterior inference in a typical scientific 
analysis, where the model 
is expensive to evaluate, is nonlinear in its model parameters, and we have numerous nuisance 
parameters we want to marginalize over. Here we give 
an example from our own research in cosmology, which was the original motivation for 
this work, because MCMC was failing to converge for this problem. We observe about 
$10^6$ galaxy positions, measured out to about half of the lookback time of the universe and distributed over a 
quarter of the sky, with the radial position determined by their redshift extracted from galaxy spectroscopic emission lines. 
Galaxy clustering is anisotropic because of the 
redshift space distortions (RSD), 
generated by the Doppler shifts proportional to the galaxy velocities. 
We can summarize the anisotropic clustering by measuring the power spectrum 
as a function of the angle $\mu$ between the line of sight direction and the wavevector of the Fourier mode. 
In this specific case we are given measured summary statistics of galaxy clustering $\hat{P}_{l}(k)$, where $l=0,2,4$
are the angular multipoles (Legendre transforming the angular dependence on $\mu$) of the power spectrum and $k$ is the wavevector amplitude. We have a model prediction 
for the summary statistic $P_{l}(k)$ that depends on 13 different parameters, of which 3 are of 
cosmological interest, since they inform us of the content of the universe, including 
dark matter and dark energy. 
Others can be viewed as nuisance parameters, although they can also be of 
interest on their own \citep{HandSeljakEtAl17}. 
We are also given the 
covariance matrix of the summary statistics (generalized 
noise matrix).  
The covariance matrix depends on the signal $P_{l}(k)$, so the derivative of the noise matrix with respect to the 
parameters needs to be included in the analysis.
We assume flat prior on the parameters and we use Gauss-Newton approximation for 
the Hessian, so in terms of equation \ref{ls} we ignore the prior term with $\bi{S}$, 
while the parameter dependence is both in $\bi{f}$ and in $\bi{N}$. 

A common complication for scientific analyses  is that the gradients are often not available in an analytic form: 
the models are evaluated as a numerical evaluation of ODEs or PDEs with many 
time steps to evolve the system from its initial conditions to the final output. 
Doing back-propagation on ODEs or PDEs with a large number of steps 
can be expensive and requires dedicated codes, 
which are often not available. In our application, we were able to do analytic 
derivatives for 9 parameters, leaving 4 to numerical finite difference method. 
We have implemented these with one sided derivatives (step size of 5\%), 
meaning that we need 5 function calls to get the function and the 
gradient. Due to the use of 
Gauss-Newton approximation we also get an approximate Hessian at no extra cost. This 
finite difference approach could be improved, for example by using some more global interpolation schemes, but for this paper we will not attempt to do so. 

Our specific optimization approach was to use L-BFGS (with L=5) for initial steps, 
switching to Gauss-Newton optimization at later steps closer to the solution. 
We started with assuming $q$ is a delta function (MAP approach), switching to 
sampling from $q$ as we approach the minimum, and gradually increasing the number
of samples $N_k$ once we are past the burn-in, reusing samples from the previous iterations after the burn-in. Here 
burn-in is defined in terms of EL$_2$O not rapidly 
changing anymore. 
Overall it took 25 iteration steps to converge
to the full non-Gaussian posterior solution. Number of samples and iteration steps used in all numerical examples are outlined in Table \ref{table1}.

Results are shown in figure \ref{fig:Pk2}. In the top 
panel the parameters are $f \sigma_8$ (product of the growth rate $f$ and the amplitude of matter fluctuations $\sigma_8$), $b_1$ (linear bias), $\sigma_c$ (velocity dispersion for central galaxies), and $f_{1h,s_Bs_B}$ (normalization parameter of the 1-halo amplitude).
It is
of interest to explore how it compares to MAP+Laplace (using Hessian of $\mathcal{L}_p$ at MAP 
to determine the inverse covariance matrix) in situations where the posterior is 
approximately Gaussian, and we show these results as well
in the 
top panel of figure \ref{fig:Pk2}. We see that MAP+Laplace can fail in the mean, or in the 
covariance matrix. This could be caused by 
the marginalization over non-Gaussian probability 
distributions of other parameters, or caused by small scale noise in the log 
posterior close to the minimum, which 
EL$_2$O improves on by averaging over several samples. 
The results have converged
to the correct posterior after 25 iterations, at which point the 
EL$_2$O value is stable and around 
0.18, which we have argued is low enough for the posteriors to be accurate. 
Here we compare to MCMC emcee package \citep{ForemanHoggEtAl13}, which initially did not converge, so we restarted 
it at the EL$_2$O best fit parameters (results are shown 
with $10^5$ samples after burn-in). 

In the bottom 
panel we explore parameters that have the most non-Gaussian posteriors;
these parameters are $f_s$ (satellite fraction), $f_{sB}$ (type B satellite fraction), both of which have positivity constraint, $b_{1,\ c_A}$ (linear bias of the type A central galaxies), and $\gamma_{b_1 s_A}$ (slope parameter in the relation between $b_{1,\ s_A}$ and $b_{1,\ c_A}$). In all cases the EL$_2$O posteriors agree
remarkably well with MCMC.  This is even the case
for the parameter $f_{sB}$, which has a positivity constraint $f_{sB}>0$, but 
is poorly constrained, with a very non-Gaussian posterior that peaks at 0. Even for 
this parameter the median and 2.5\%, 97.5\% lower and upper limits agree with MCMC. When we model this parameter 
with the unconstrained transformation method, we see that the probability rapidly 
descends to zero at the boundary $f_{sB}>0$. The 
reflective boundary method corrects this and gives a 
better result at the boundary, 
as also shown in the same figure. In this example, with reflective boundary, 
we allowed the posterior to 
go to -0.2 on this parameter. 
We do not show MAP+Laplace results since 
they poorly match these non-Gaussian 
posteriors. 

\setlength\extrarowheight{3pt}
\begin{table}[t]
\centering
\begin{tabular}{l | c c c c}
\hline
\hline 
& \phantom{...} Ex. 4.1\phantom{...} & \phantom{...}4.2\phantom{...} & \phantom{.....}4.3 \phantom{...}& \phantom{...}4.4\phantom{...}  \\
\hline
$N_k$ & 1-10 & 1-10 & 1-10 & 1-10 \\ 
$N_{\mathrm{iteration}}$ & 10 & 15 & 15 & 25 \\
$N_{\rm tot}$ & 25 & 25 & 25 & 125 \\
\hline
\end{tabular}
\caption{Number of samples per iteration $N_k$, number of iteration steps $N_{\mathrm{iteration}}$, 
and total number of $\tilde{\mathcal{L}}_p$ evaluations (incldding 
burn-in) for 4 different numerical examples presented in this work. Example 4.4 does not have analytic gradients for 4 parameters, which are evaluated with a finite difference instead.}
\label{table1}
\end{table}

\begin{figure}[t!]
\centering \includegraphics[height=0.39\textwidth]{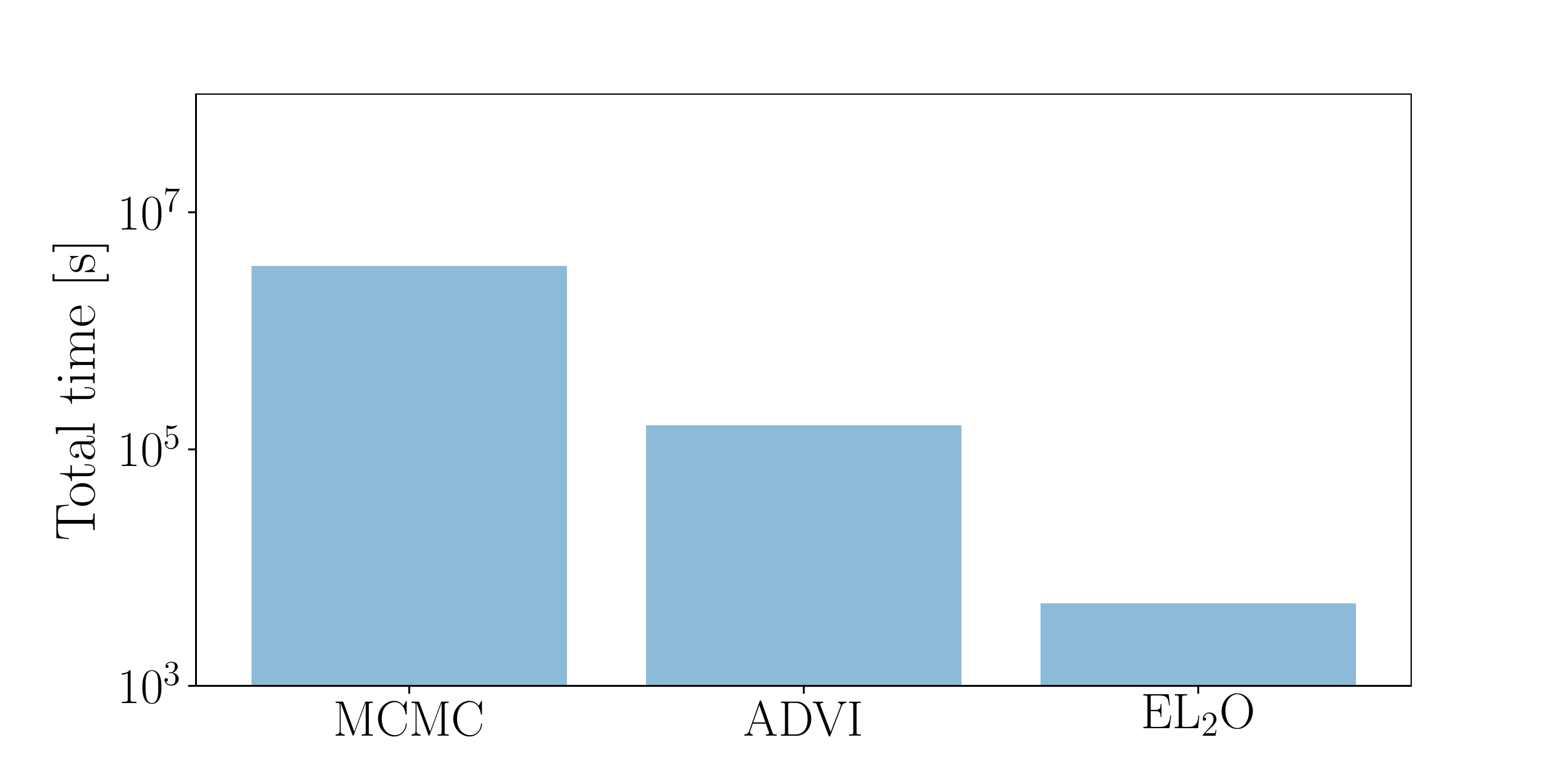}
\label{fig:timing}
\caption{Timing results for 13-dimensional example of section 4.4. EL$_2$O is about 1000 times faster than MCMC and 200 times 
faster than ADVI using the same parametrization (but which did not converge to exact posteriors, as seen in figure \ref{fig:Pk2}).}
\end{figure}

\section{Discussion and conclusions}
\label{sec5}

The main goal of this work is to develop a method that 
gives reliable and smooth parameter posteriors, 
while also 
minimizing the number of 
calls to log joint probability $\tilde{\mathcal{L}}_p$. In many settings, specially for scientific 
applications, an 
evaluation of $\tilde{\mathcal{L}}_p$ can be extremely costly. 
The current gold standard are MCMC methods, 
which asymptotically converge to the correct answer, but require a very large 
number of likelihood evaluations, often exceeding $10^5$ or more. 
Scientific models are becoming more and more 
sophisticated, which comes at a heavy computational cost in terms
of evaluation of $\tilde{\mathcal{L}}_p$.   
Using brute force MCMC sampling methods in these situations
is practically impossible.  
In this paper we follow the optimization approaches of MAP+Laplace and stochastic VI \citep{KucukelbirTRGB17}, 
but we modify and extend these in several directions. 
The focus of this paper is 
on low dimensionality problems, where doing matrix inversion and Cholesky 
decomposition of the Hessian is not costly compared to evaluating 
$\tilde{\mathcal{L}}_p$. In practice this limits the method to of order thousands of 
parameters, if they are correlated so that the full rank matrix description is 
needed. If one can adopt sparse matrix approximations one can increase the 
dimensionality of the problem. 

Both VI and MCMC methods rely
on minimizing KL divergence. 
When the problem is not 
tractable using deterministic methods this minimization uses sampling, and 
this leads to a sampling noise in optimization 
that is only reduced as inverse square root 
of the number of samples. This can be traced to the feature of KL divergence that 
its integrand does not have 
to be positive, even if KL divergence is. minimization of
KL divergence only makes sense in the context of the KL divergence integral $\int dz q (\ln q -\ln p)$
: it is only positive after the integration. Deterministic integration is only feasible in very low dimensions, and stochastic integration via Monte Carlo converges slowly, as $N_k^{-1/2}$.
In this 
paper we propose instead to minimize ${\rm EL_2O}$,
Euclidean L$_2$ distance squared between the log posterior $\mathcal{L}_p$, which 
we evaluate as $\tilde{\mathcal{L}}_p+\ln \bar{p}$, where $\ln \bar{p}$ is an 
unknown constant, 
and the equivalent terms of its 
approximation $\mathcal{L}_q$. EL$_2$O is based on comparing $\ln q(z_k)$ and $\ln p(z_k)$ at the same sampling 
points $z_k$: if the two distributions are to be equal they should agree at every sampling point, up to the 
normalization constant. There is no need 
to perform the integral to obtain a useful minimization procedure and there is no stochastic integration noise, 
only one extra optimization parameter due to the unknown normalization. 
When available 
we add its higher order 
derivatives, where we do not have to distinguish between 
$\mathcal{L}_p$ and $\tilde{\mathcal{L}}_p$. This is 
evaluated as expectation over some approximate probability distribution $\tilde{p}$ close to $p$. 
While one can construct many different f-divergences, 
${\rm EL_2O}$ optimization 
agrees with KL divergence minimization based VI 
in the high sampling limit, if samples are generated from $\tilde{p}=q$. 
However, for a finite number of samples 
the resulting algorithm differs from 
recent VI methods such as ADVI \citep{KucukelbirTRGB17}, or the response method \citep{GiordanoBJ18}.
While $t\ln t$ (KL divergence) and $t\ln^2t$ f-divergence minimization (EL$_2$O) seem very similar, they are fundamentally 
different, the former more related to stochastic integration. 
 
A first advantage of ${\rm EL_2O}$ is that it 
has no sampling noise if the family of models $q$ covers $p$, in contrast to the stochastic minimization of KL divergence, as we demonstrate 
in section \ref{sec2}. 
In this case
the method gives exact solution and  ${\rm EL_2O}=0$, 
as long as we have enough constraints as the parameters, 
and it does not even matter where the samples are drawn from.  In this limit additional samples make the problem over-constrained, 
 which does not improve the result. 
If the family of $q$ is too simple to cover $p$ 
then the 
results fluctuate depending on the drawn samples and the 
convergence is slower. Having more 
expressive $q$ so that it is closer to $p$ 
makes the convergence faster even if there are 
more parameters to be determined. 
 In practical examples 
 we have observed this behavior once  ${\rm EL_2O}$ dropped 
 below 0.2 (e.g. figure \ref{ELO}), where we approach 
 exact inference. This property of 
 ${\rm EL_2O}$ is different from a stochastic KL divergence minimization, 
 which is noisy and will typically take longer to converge 
 as the number of paramaters increases. 
Moreover, as we argue in section \ref{sec2}, even in the 
simplest setting of Gaussian posterior stochastic 
KL divergence minimization is not a convex problem for 
a finite number of samples, while EL$_2$O is. In this 
setting EL$_2$O is not only convex, but also linear, 
so normal equations (or a single Newton update) give the complete solution. 
While we use L$_2$ distance in this paper, L$_1$ distance differs
from KL divergence only in taking the absolute 
value of the log posterior difference, and also 
has no sampling noise. In terms of f-divergence 
${\rm EL_2}$ corresponds to f-divergence $t(\ln t)^2$ and 
${\rm EL_1}$ to $t|\ln t|$, in contrast to $t\ln t$ for KL divergence. 


Second advantage of ${\rm EL_2O}$ is that its value can be used to quantify the quality of the
solution: when it is 
small (less than 0.2 for our examples) $p$ is well described with $q$ and 
we may exit optimization. 
In contrast, in KL divergence based VI the value of the lower bound (ELBO) does not have an absolute 
meaning since it is related to the normalization $\ln p(x)$ (free energy bound, \cite{JordanEtAl99}): while relative changes of ELBO are meaningful, the absolute value is not and other methods are needed to assess the quality of the answer \citep{YaoVSG18}. Even though we do not need it
explicitly, EL$_2$O can easily optimize on and output an 
approximation to $\ln p(x)$. This can be useful for evidence or
Bayes factor evaluations, which we will pursue 
elsewhere. 
Sometimes we find low EL$_2$O already for a
full rank Gaussian and sometimes we need to go beyond it.
There is not much computational 
benefit in using $q$ that is simpler than a full rank Gaussian, 
as long as Cholesky decomposition and matrix inversion
are not a computational bottleneck: only the means are being 
optimized and not the covariance matrix elements. 
${\rm EL_2O}$ value provides a 
diagnostic to asses the quality of the approximate posterior, so when it is large 
one can extend the family of models $q$
to remedy the situation. The strategy we advocate is to improve $q$ until a low 
value of ${\rm EL_2O}$ is reached.
However, 
 in contrast to recent trends in machine learning with 
 many layers of nonlinear transformations (e.g. normalizing 
 flows, \cite{RezendeMohamed15}), we advocate a single 
 transformation with few parameters only, 
 such that the number 
 of $\tilde{\mathcal{L}}_p$ evaluations is minimized, 
 and the analytic marginalization remains possible. 
 If this fails to reduce EL$_2$O one can improve the approximate posterior 
 by adding one
 Gaussian mixture or non-bijective transformation at a 
 time. 

Third advantage of ${\rm EL_2O}$ is its flexibility in choosing the sampling 
proposal $\tilde{p}$, which distinguishes it from the KL divergence based VI, 
which can only sample 
from $q$, and requires reparametrization trick for optimization \citep{KingmaWelling13,RezendeMW14}. This trick is not needed for ${\rm EL_2O}$ optimization.
For $\tilde{p}$ one can use $q$, and iterate on it, which gives results 
identical to VI in the large sampling density limit. 
But we can also use true $p$ if we
have a way to evaluate its samples, which is not only possible 
but easier than sampling from $q$ for forward model problems. 
We can also sample from both $p$ and $q$, 
which may avoid some of the pitfalls of the other sampling 
approaches. 

Because we can choose different sampling proposals we 
can also address the quality of the results as a 
function of this choice. 
In the forward model 
example where sampling from $p$ is easy we 
have found that for simple $q$ (with ${\rm EL_2O} > 0.4$) 
there was a difference between sampling from $q$ 
versus sampling from 
$p$, the latter giving overall better results. 
These differences were reduced as we increased the 
expressivity of $q$, for example when going from FR Gaussian to NL+FR, consistent with the statement above that more 
expressive $q$ improves the results and in the limit of very expressive 
$q$ we approach exact inference. Samples from $q$ and 
samples from $p$ gave nearly the same EL$_2$O value for the same $\mathcal{L}_q$, 
suggesting that sampling from 
$q$ may not be a fundamental limitation of EL$_2$O divergence based
variational methods, 
but more a limitation of using 
insufficiently expressive forms
of $q$. Recently, several divergences have been introduced 
(e.g. \cite{RanganathTAB16,DiengTRPB17})
to counter these suggested problems of KL divergence, but here
we argue that with 
sufficiently expressive $q$ this may not be an issue 
for EL$_2$O method. 
We also do not observe that in EL$_2$O 
sampling from $q$ leads to 
a significantly narrower approximation than sampling from $p$ once we 
go to NL+FR for $q$, in contrast 
to the arguments in the context of FRVI \citep{Bishop07}.
In this example 
further improvement when 
sampling from $p+q$ was negligible. 
While this is based on a limited set of examples
and
deserves further study, we expect that with sufficiently 
expressive $q$ EL$_2$O value can be driven to zero and we approach 
exact inference, so that for most problems 
we only need to 
consider sampling from $\tilde{p}=q$. 



Fourth advantage of ${\rm EL_2O}$ is its ability to use gradient and Hessian information (and even higher 
order derivatives if available), while preserving its sampling noise free nature. 
A significant trend in recent years, in machine learning, statistics and 
scientific computing,
has been the development of analytic gradients and Hessians of $\tilde{\mathcal{L}}_p$ using 
methods such as backpropagation. 
${\rm EL_2O}$ can  
take advantage of this and we present both the gradient based version  and the gradient and Hessian based version. When ${\rm EL_2O}$ is using only the gradient information it can be 
related to Fisher divergence \citep{Hammad78}. 
The Hessian version converges especially rapidly, as every sample gives $M(M+3)/2+1$ 
constraints for $M$ dimensions, enough to fit a full rank 
Gaussian component in a Gaussian mixture model. Moreover, 
no optimization is needed to 
determine the covariance matrix of the Gaussian $q$, as its inverse is simply given by averaging 
the Hessian over the samples. When Hessian is not available 
one can use the gradient information as in equation \ref{gradsig} to achieve the same. 
For nonlinear least squares problems Hessian in the 
Gauss-Newton approximation can be obtained at the same cost as the gradient, and we found this approximation to give reliable 
posteriors in a realistic scientific application. For harder problems, 
where MAP, 
MFVI and FRVI with Gaussian $q$ all fail, we found the 
Gaussian mixture model to work well. 
In our applications 
we combine full rank Gaussian mixtures model 
with one dimensional nonlinear transforms to fit a
general posterior, while at the same time also being able to do analytic 
marginalization over any parameters. 

In most applications the number of iterations was 
comparable to the number of likelihood evaluations: we start
with a single value for the mean 
MAP strategy for the burn in, then slowly increase the 
number of samples we average over by reusing samples from 
past iterations, typically to about 10-15 samples, if 
gradient and Hessian are available. 
In a realistic scientific
application in the field of cosmology, 
with 13 parameters and no analytic derivatives for 4 of them, we obtained good 
posteriors with about 25 iterations, with 5 calls each to obtain the finite difference gradients, as compared to $10^5$ iterations for MCMC. 
This was a particularly difficult problem for MCMC, 
which did not even converge until we restarted it at 
${\rm EL_2O}$ solution, and this problem was the original motivation for this work. Using the same parametrization on
ADVI, with $2\times 10^4$ calls, we obtained
worse posteriors despite 200 times higher computational cost, a consequence of noise in KL divergence it is minimizing. 
For many of the parameters the posteriors are 
very non-Gaussian, and we found remarkable agreement between 
${\rm EL_2O}$ and MCMC using full rank Gaussian and a nonlinear 
transformation with two or three parameters for each dimension. 

In this 
example evaluation of $10^5$ samples was feasible and we were able to 
compare the results to MCMC, but in many realistic 
situations MCMC would not even be feasible, and methods such as ${\rm EL_2O}$ may be 
one of the few possible alternatives. 
More generally, given the ubiquitousness of KL divergence in many 
applications in statistics and machine learning, 
${\rm EL_2O}$ may also be useful as an alternative to KL divergence beyond the
posterior inference applications described in this paper. 
One important application where EL$_2$O has 
an advantage over MCMC is calculation of normalization or
evidence $p(x)$. As discussed in this paper, 
one of the optimization 
outputs of EL$_2$O procedure is $\ln \tilde p$, 
an approximation to 
$\ln p(x)$, which can in turn be used for
model comparison by comparing the evidences 
between the
different models. 

\section{Acknowledgments}
We thank Andrew Charman, Yu Feng, Ryan Giordano, He Jia, Francois Lanusse, Patrick McDonald, Jeffrey Regier and 
Matias Zaldarriaga for useful discussions. US is supported by grants NASA NNX15AL17G, 80NSSC18K1274, NSF 1814370, and NSF 1839217.

\bibliography{cosmo}

\end{document}